\documentclass[11pt]{article}
\usepackage[preprint]{acl}
\usepackage{times}
\usepackage{latexsym}
\usepackage[T1]{fontenc}
\usepackage[utf8]{inputenc}
\usepackage{microtype}
\usepackage{inconsolata}
\usepackage{booktabs}
\usepackage{array}
\usepackage{graphicx}
\usepackage{pifont}
\usepackage{amsmath}
\usepackage{amssymb}
\usepackage{enumitem}
\usepackage{tikz}
\usetikzlibrary{arrows.meta}
\graphicspath{{figures/}}
\title{Less but Sufficient: Relation Knowledge Selects Compact\\Knowledge-Graph Contexts, and What External Checks\\Can and Cannot Catch}

\author{Seunghui Jwa$^{1}$ \quad Minsu Oh$^{1}$ \quad Sangwoo Hong$^{2}$ \quad Yeo Chan Yoon$^{1}$\thanks{Corresponding author.}\\
\textnormal{$^{1}$Department of Artificial Intelligence, Jeju National University}\\
\textnormal{$^{2}$Department of Computer Science and Engineering, Konkuk University}}

\hypersetup{
  pdftitle={Less but Sufficient: Relation Knowledge Selects Compact Knowledge-Graph Contexts, and What External Checks Can and Cannot Catch},
  pdfauthor={Seunghui Jwa, Minsu Oh, Sangwoo Hong, Yeo Chan Yoon}
}

\newcommand{\tabinputs}{%
\begin{table*}[t]
\centering
\footnotesize
\setlength{\tabcolsep}{2.2pt}
\resizebox{\ifdim\width>\textwidth \textwidth\else \width\fi}{!}{%
\begin{tabular}{l c cccr cccr cccr cccr}
\toprule
& & \multicolumn{4}{c}{\textbf{PathQuestion}} & \multicolumn{4}{c}{\textbf{MetaQA}} & \multicolumn{4}{c}{\textbf{WebQSP}} & \multicolumn{4}{c}{\textbf{CWQ}} \\
\cmidrule(lr){3-6}\cmidrule(lr){7-10}\cmidrule(lr){11-14}\cmidrule(lr){15-18}
\textbf{context} & \textbf{CLUTRR} & \textbf{exact} & \textbf{Hit} & \textbf{F1} & \textbf{tok.} & \textbf{exact} & \textbf{Hit} & \textbf{F1} & \textbf{tok.} & \textbf{exact} & \textbf{Hit} & \textbf{F1} & \textbf{tok.} & \textbf{exact} & \textbf{Hit} & \textbf{F1} & \textbf{tok.} \\
\midrule
all facts or neighbourhood & 53.0 & 50.2 & 70.2 & 62.2 & 1,073 & 47.3 & 75.2 & 59.3 & 3,630 & 24.8$^\dagger$ & 51.6$^\dagger$ & 39.4$^\dagger$ & 6,557$^\dagger$ & -- & -- & -- & -- \\
similarity top-$k$ & 43.2 & 33.0 & 51.9 & 44.8 & 191 & 26.0 & 56.5 & 39.6 & 375 & 21.6 & 46.9 & 35.3 & 424 & -- & -- & -- & -- \\
KAPING & 33.7 & 19.0 & 33.4 & 26.2 & 172 & 31.3 & 50.9 & 38.7 & 224 & 28.5 & 52.6 & 41.6 & 311 & 10.4 & 23.7 & 17.7 & 318 \\
SubgraphRAG & -- & 52.4 & 70.7 & 63.5 & 2,151 & 49.7 & \textbf{86.0} & 66.9 & 1,020 & 43.9 & \textbf{78.6} & 64.5 & 2,139 & 27.0 & \textbf{54.3} & \textbf{41.5} & 2,067 \\
Think-on-Graph & -- & 44.5 & 65.8 & 55.8 & 3,658 & 33.5 & 59.8 & 44.4 & 2,906 & 29.0 & 63.3 & 46.5 & 5,676 & -- & -- & -- & -- \\
RoG & -- & -- & -- & -- & -- & -- & -- & -- & -- & 37.5 & 76.6 & 61.7 & 516 & 22.4 & 48.6 & 36.7 & 653 \\
selected compact context (ours) & \textbf{56.3} & \textbf{72.4} & \textbf{87.7} & \textbf{82.2} & 191 & \textbf{59.5} & 79.8 & \textbf{71.1} & 345 & \textbf{54.1} & 77.0 & \textbf{67.9} & 414 & \textbf{27.3} & 52.0 & 40.7 & 505 \\
\midrule
annotated path (reference) & 55.4 & 59.3 & 90.4 & 74.5 & 167 & 72.4 & 96.6 & 87.9 & 383 & 64.6 & 87.1 & 79.1 & 425 & 35.2 & 64.4 & 51.7 & 468 \\
\bottomrule
\end{tabular}
}%
\caption{LLM answers per context, forced-choice prompt, mean over the seven open LLMs (\%). A published method's row is the context it selects, not its full pipeline; neighbourhood: the one-hop neighbourhood of the question's entities. $^\dagger$Six LLMs, as WebQSP's neighbourhoods exceed Gemma-3-27B's window. Exact: the whole gold set (CLUTRR: the label); Hit: some gold answer; tok.: prompt tokens per question, summed over a method's LLM calls (on CLUTRR 161 to 187 for every context). Similarity top-$k$ holds as many facts as ours (on MetaQA as the classifier's context, 14.6 facts against the lexicon's 12.6). Bold: the best context, excluding the annotated path, a reference that misses PathQuestion's further answers and is empty for 125 WebQSP and 787 CWQ questions. Ours (on MetaQA the lexicon's context) is read on every question (pass rates: Table~\ref{tab:gates}). --: not run; CLUTRR's stories are given whole, RoG released only WebQSP and CWQ paths, and the other CWQ cells are in Appendix~\ref{app:cwq}.}
\label{tab:inputs}
\end{table*}
}

\newcommand{\tabfaith}{%
\begin{table*}[t]
\centering
\footnotesize
\setlength{\tabcolsep}{5pt}
\begin{tabular}{l r r r r r r r}
\toprule
\textbf{pipeline} & \textbf{Ll-3B} & \textbf{Qw-7B} & \textbf{Ll-8B} & \textbf{Ex-7.8B} & \textbf{Ge-27B} & \textbf{Qw-32B} & \textbf{Ex-32B} \\
\midrule
\multicolumn{8}{l}{\emph{Balanced score, one necessary fact deleted (usually disconnects the queried people)}} \\
LLM decides alone & 0.303 & 0.644 & 0.386 & 0.399 & 0.770 & 0.715 & 0.747 \\
gate: queried people connected & 0.593 & 0.661 & 0.686 & 0.680 & 0.778 & 0.831 & 0.846 \\
gate: derivability, rules & 0.601 & 0.668 & 0.693 & 0.687 & 0.784 & 0.838 & 0.853 \\
\midrule
\multicolumn{8}{l}{\emph{Three-way score, one false fact added as well}} \\
LLM decides alone & 0.208 & 0.577 & 0.259 & 0.269 & 0.573 & 0.477 & 0.509 \\
gates: derivability and consistency, rules & 0.733 & 0.778 & 0.795 & 0.791 & 0.856 & 0.892 & 0.902 \\
gates: learned, GRU derivability, count consistency & 0.709 & 0.762 & 0.770 & 0.767 & 0.835 & 0.867 & 0.877 \\
gates: learned, GRU added to consistency & 0.729 & 0.775 & 0.791 & 0.787 & 0.852 & 0.888 & 0.897 \\
\bottomrule
\end{tabular}
\caption{Faithful abstention on CLUTRR, abstention prompt (scores: Section~\ref{sec:checks}; three-way: with one reversed-edge or far contradiction). Policies: the LLM alone, or the LLM answering where the named checks pass and abstaining where one fails, which is how the contradiction experiments use the consistency check; no derived answer is returned. With full detection a gated score is (accuracy + 1)/2 or (accuracy + 2)/3, from the LLM's accuracy where the checks pass. Learned: Appendix~\ref{app:learned}, with the GRU as derivability check in both rows; with the GRU added, the consistency check flags exactly the same compact contexts as one whose GRU replaces the counts. Ll: Llama, Qw-7B: Qwen2.5-7B, Qw-32B: Qwen3-32B, Ex: EXAONE, Ge: Gemma; billions of parameters.}
\label{tab:faith}
\end{table*}
}

\newcommand{\tablearned}{%
\begin{table}[t]
\centering
\footnotesize
\setlength{\tabcolsep}{2.4pt}
\resizebox{\ifdim\width>\columnwidth \columnwidth\else \width\fi}{!}{%
\begin{tabular}{l r r r}
\toprule
\textbf{contexts} & \textbf{counts} & \textbf{+GRU} & \textbf{rules} \\
\midrule
\multicolumn{4}{l}{\emph{Contradicted compact contexts: share detected}} \\
CLUTRR, reversed edge / far & 99.9 / 87.6 & 99.9 / 99.9 & 99.9 / 99.8 \\
PathQuestion, reversed / two-step & 100.0 / 100.0 & 100.0 / 100.0 & -- \\
WebQSP, reversed / via a record & 96.8 / 83.0 & 96.8 / 83.0 & -- \\
WebQSP, second value & 96.4 & 96.4 & -- \\
\midrule
\multicolumn{4}{l}{\emph{False fact in the graph before selection: share detected}} \\
CLUTRR, shortcut / substitution & 0.0 / 2.0 & 0.0 / 2.1 & 0.0 / 2.1 \\
PathQuestion, substitution & 0.0 & 0.0 & -- \\
WebQSP, substitution / second value & 0.2 / 63.9 & 0.2 / 63.9 & -- \\
\midrule
\multicolumn{4}{l}{\emph{Consistent contexts: false alarms}} \\
CLUTRR, compact / story & 0.0 / 0.1 & 0.0 / 0.1 & 0.0 / 0.8 \\
PathQuestion, compact / 1-hop & 0.0 / 1.8 & 0.1 / 1.8 & -- \\
WebQSP, compact / SubgraphRAG's & 0.8 / 0.9 & 0.9 / 1.8 & -- \\
\bottomrule
\end{tabular}
}%
\caption{Contexts flagged (\%; higher is better for contradicted contexts, lower for consistent ones), thresholds giving 1\% false alarms on consistent calibration contexts. Counts: the default learned check (Appendix~\ref{app:learned}); +GRU: the path classifier added; before selection: false facts that are selected, pass the derivability check and are flagged, as a share of the n entered (WebQSP substitution: 2 of 1,132; 14 counting contexts that fail the check, Appendix~\ref{app:checkable}); TransE plausibility alone detects at most 1.2\% of the PathQuestion and WebQSP probes; --: no rules; via a record: through an unnamed record. WebQSP: 126, 53 and 83 contradicted contexts, 1,132 and 83 false facts in the graph. The injected contradictions create the second path the path scores need, which 2.5\% of the named fact mentions in clean compact contexts have (Table~\ref{tab:checkable}). MetaQA's consistency is not tested, as all its relations lead from a film to an attribute.}
\label{tab:learned}
\end{table}
}

\newcommand{\tabdecomp}{%
\begin{table}[t]
\centering
\scriptsize
\setlength{\tabcolsep}{3pt}
\resizebox{\ifdim\width>\columnwidth \columnwidth\else \width\fi}{!}{%
\begin{tabular}{p{46mm} r r}
\toprule
\textbf{policy} & \textbf{annotated} & \textbf{weak} \\
\midrule
LLM on SubgraphRAG's 100 facts (no selection, no gate) & 43.9 & 43.9 \\
LLM on the compact context (selection) & 54.1 & 47.4 \\
derived answer where the routing gate passes, else LLM on the compact context (routing) & 59.1 & 47.8 \\
derived answer where the routing gate passes, else LLM on SubgraphRAG's facts & 59.3 & 46.9 \\
LLM on the compact context where the routing gate passes, else on SubgraphRAG's facts & 54.3 & 46.6 \\
\midrule
\multicolumn{3}{@{}p{\dimexpr\columnwidth-2\tabcolsep\relax}@{}}{routing gate instead of the vote share alone, as many derived answers returned: wrong derived answers 35.4 to 34.9\%} \\
\multicolumn{3}{@{}p{\dimexpr\columnwidth-2\tabcolsep\relax}@{}}{consistency check added to routing: $-$0.1 to +0.0 per LLM; 12 of 1,385 passing contexts flagged} \\
\bottomrule
\end{tabular}
}%
\caption{What each part contributes on WebQSP, chain predictor fixed: mean exact match of the seven open LLMs (\%) per policy, with annotated and with weakly supervised chains, that is, shortest question-to-answer paths with their own check as the routing gate. Below, the routing gate against the vote share alone at equal coverage, and the consistency check added to routing as a variant. Per LLM: Appendices~\ref{app:passfail}, \ref{app:routing} and~\ref{app:layers}.}
\label{tab:decomp}
\end{table}
}

\newcommand{\tabanswer}{%
\begin{table}[t]
\centering
\footnotesize
\setlength{\tabcolsep}{1.5pt}
\resizebox{\ifdim\width>\columnwidth \columnwidth\else \width\fi}{!}{%
\begin{tabular}{l r r r r r r}
\toprule
& & \multicolumn{3}{c}{\textbf{routing gate passes}} & \multicolumn{2}{c}{\textbf{all questions}} \\
\cmidrule(lr){3-5}\cmidrule(lr){6-7}
\textbf{knowledge} & \textbf{passes} & \textbf{derived} & \textbf{\begin{tabular}[b]{@{}c@{}}best\\LLM\end{tabular}} & \textbf{GPT-5} & \textbf{derived} & \textbf{routed} \\
\midrule
CLUTRR, rules & 99.8 & 99.4 & 73.3 & 91.7$^\ast$ & 99.2 & 99.2--99.3 \\
MetaQA, classifier & 100.0 & 99.9 & 89.0 & 97.0 & 99.9 & 99.9 \\
MetaQA, lexicon & 77.1 & 99.6 & 90.6 & 98.0 & 76.8 & 76.9--77.4 \\
PathQuestion, lexicon & 77.0 & 96.2 & 86.2 & 91.9 & 89.6 & 86.3--93.6 \\
WebQSP, chains & 85.1 & 65.1 & 66.0 & 69.8 & 56.3 & 57.8--60.9 \\
CWQ, chains & 24.5 & 68.6 & 67.9 & 70.3 & 22.3 & 24.4--38.4 \\
\bottomrule
\end{tabular}
}%
\caption{Who should answer: share of questions passing the routing gate of Section~\ref{sec:method} (\%); accuracy there of the derived answer, the best open LLM and GPT-5 with reasoning; and, on all questions, the derived answer alone and routed, that is, the derived answer where the routing gate passes, else the open LLM, with no consistency check, as a range over the seven LLMs. With the check instead, WebQSP's derived answer is 78.7 and GPT-5 82.2, CWQ's 52.0 and 70.5 (other CWQ LLMs: Appendix~\ref{app:cwq}). $^\ast$CLUTRR's 264-question subset, where the derived answer is 99.2.}
\label{tab:answer}
\end{table}
}

\newcommand{\tabprecgate}{%
\begin{table}[t]
\centering
\scriptsize
\setlength{\tabcolsep}{3pt}
\begin{tabular}{@{}l r l r r l@{}}
\toprule
& \multicolumn{2}{c}{\textbf{test}} & \multicolumn{3}{c}{\textbf{held-out}} \\
\cmidrule(lr){2-3}\cmidrule(lr){4-6}
\textbf{benchmark} & \textbf{cov.} & \textbf{prec.} & \textbf{cov.} & \textbf{right / ret.} & \textbf{prec.} \\
\midrule
WebQSP & 19.0 & 91.3 [87.6, 93.9] & 17.6 & 79 / 88 & 89.8 [81.7, 94.5] \\
CWQ & 5.5 & 96.4 [93.1, 98.9] & 4.9 & 70 / 73 & 95.9 [87.0, 100.0] \\
\bottomrule
\end{tabular}
\caption{The precision gate, the derivability check passing with one derived answer: share of questions answered (cov., \%) and share of those answers that are exact (prec., \%) on the test sets and in the held-out test specified in advance, with its right and returned counts. Brackets: 95 percent intervals, Wilson on WebQSP, seed-cluster bootstrap on CWQ.}
\label{tab:precgate}
\end{table}
}

\newcommand{\tabgates}{%
\begin{table}[t]
\centering
\footnotesize
\setlength{\tabcolsep}{3pt}
\resizebox{\ifdim\width>\columnwidth \columnwidth\else \width\fi}{!}{%
\begin{tabular}{@{}l r r r r@{}}
\toprule
& \multicolumn{2}{c}{\textbf{test}} & \multicolumn{2}{c}{\textbf{held-out}} \\
\cmidrule(lr){2-3}\cmidrule(lr){4-5}
\textbf{gate and benchmark} & \textbf{cov.} & \textbf{prec.} & \textbf{cov.} & \textbf{prec.} \\
\midrule
\multicolumn{5}{@{}l}{\emph{derivability check}: an answer follows from the context} \\
\quad WebQSP & 45.6 & 78.7 & 44.4 & 81.5 \\
\quad CWQ & 11.8 & 52.0 & 9.7 & 62.1 \\
\multicolumn{5}{@{}l}{\emph{routing gate}: where the derived answer replaces the LLM's$^a$} \\
\quad WebQSP: the check or $\leq$ 20 answers & 85.1 & 65.1 & 85.4 & 66.0 \\
\quad CWQ: one derived answer & 24.5 & 68.6 & 22.6 & 69.0 \\
\quad PathQuestion: the check & 77.0 & 96.2 & -- & -- \\
\quad MetaQA, lexicon: the check & 77.1 & 99.6 & -- & -- \\
\multicolumn{5}{@{}l}{\emph{precision gate}, derived answers alone: the check and one answer$^b$} \\
\quad WebQSP & 19.0 & 91.3 & 17.6 & 89.8 \\
\quad CWQ & 5.5 & 96.4 & 4.9 & 95.9 \\
\bottomrule
\end{tabular}
}%
\caption{The derivability check and the two gates built on it: share of questions passing (cov.) and share of these whose derived answer is exact (prec.), on the test questions and on the held-out questions of the pre-specified plan (\%; Section~\ref{sec:checks}). $^a$Picked on validation questions by a procedure fixed in advance among 43 rules defined after test comparisons; no LLM read PathQuestion or MetaQA validation questions, so their routing gate is the check. $^b$Defined after its CWQ test precision had been seen.}
\label{tab:gates}
\end{table}
}

\newcommand{\tabanswerfull}{%
\begin{table*}[t]
\centering
\footnotesize
\setlength{\tabcolsep}{3pt}
\resizebox{\ifdim\width>\textwidth \textwidth\else \width\fi}{!}{%
\begin{tabular}{l r r r r r r r r r}
\toprule
& & \multicolumn{2}{c}{\textbf{routing gate passes}} & \multicolumn{2}{c}{\textbf{routing gate fails}} & \multicolumn{4}{c}{\textbf{all questions}} \\
\cmidrule(lr){3-4}\cmidrule(lr){5-6}\cmidrule(lr){7-10}
\textbf{relation knowledge} & \textbf{passes} & \textbf{derived} & \textbf{LLMs} & \textbf{derived} & \textbf{LLMs} & \textbf{derived alone} & \textbf{routed} & \textbf{gain} & \textbf{$>$derived} \\
\midrule
CLUTRR, rules & 99.8 & 99.4 & 36.3--73.3 & -- & -- & 99.2 & 99.2--99.3 & +26.0 to +63.0 & 0/7 \\
MetaQA, classifier & 100.0 & 99.9 & 58.4--89.0 & -- & -- & 99.9 & 99.9 & +10.9 to +41.4 & 0/7 \\
MetaQA, lexicon & 77.1 & 99.6 & 64.3--90.6 & 0.0 & 0.5--2.9 & 76.8 & 76.9--77.4 & +6.9 to +27.2 & 0/7 \\
PathQuestion, lexicon & 77.0 & 96.2 & 55.6--86.2 & 67.8 & 53.3--85.0 & 89.6 & 86.3--93.6 & +7.7 to +31.2 & 3/7 \\
WebQSP, chains & 85.1 & 65.1 & 50.5--66.0 & 6.2 & 16.0--36.6 & 56.3 & 57.8--60.9 & $-$0.7 to +12.5 & 7/7 \\
WebQSP, lexicon & 25.3 & 44.2 & 41.5--48.8 & 22.0 & 21.5--32.5 & 27.6 & 27.3--35.4 & $-$1.2 to +0.7 & 5/7 \\
CWQ, chains & 24.5 & 68.6 & 32.1--67.9 & 7.3 & 10.0--28.6 & 22.3 & 24.4--38.4 & +0.2 to +8.9 & 7/7 \\
\bottomrule
\end{tabular}
}%
\caption{Who should answer, on compact contexts: questions passing the routing gate (\%; on WebQSP the check or at most 20 derived answers, on CWQ one derived answer; WebQSP, lexicon: its check); accuracy of the derived answer and the LLMs where it passes and fails (--: under 1\% fail); routed: the derived answer if it passes, else the LLM, with no consistency check; gain over the LLM alone (points); $>$derived: LLMs whose routing beats the derived answer alone (Holm-adjusted $p<0.05$).}
\label{tab:answerfull}
\end{table*}
}

\newcommand{\tabcheckable}{%
\begin{table}[t]
\centering
\scriptsize
\setlength{\tabcolsep}{3pt}
\resizebox{\ifdim\width>\columnwidth \columnwidth\else \width\fi}{!}{%
\begin{tabular}{@{}l r r r@{}}
\toprule
\multicolumn{4}{@{}l}{\emph{(a) Named fact mentions (17,881) and contexts (2,895)}} \\
 & \textbf{judgeable} & \textbf{\%} & \textbf{flagged (\%)} \\
mentions, second path & 451 & 2.5 & 27.6 \\
mentions, single-value rule only & 399 & 2.2 & 0.0 \\
mentions, either & 850 & 4.8 & 14.6 \\
contexts, second path & 107 & 3.7 & 29.9 \\
contexts, either & 406 & 14.0 & 7.9 \\
\midrule
\multicolumn{4}{@{}l}{\emph{(b) Natural errors in these contexts, by labelling}} \\
 & \textbf{consensus} & \textbf{GPT-5} & \textbf{both agree} \\
error mentions & 32 & 91 & 28 \\
\quad with path evidence (judgeable) & 2 & 7 & 0 \\
\quad flagged & 2 & 6 & 0 \\
AUC, error-centred contexts & 0.52 & 0.53 & 0.52 \\
\bottomrule
\end{tabular}
}%
\caption{What the consistency check can judge in clean compact WebQSP and CWQ test contexts. (a) Mention: one fact in one context; named: both entities named. Judgeable: a second path joins the mention's entities, or the single-value rule applies; flagged: false alarms among mentions not confirmed as errors, or contexts with a flagged mention. (b) Errors confirmed against Wikidata by two LLM passes and a tie-break (consensus), by GPT-5 and by both; suspicion-score AUC against supported facts, chance 0.5 (Appendix~\ref{app:learned}).}
\label{tab:checkable}
\end{table}
}

\begin{document}
\sloppy
\emergencystretch=3em
\maketitle

\begin{abstract}
\looseness=-1 Large language models (LLMs) read knowledge-graph facts ranked by relevance, leaving open whether they
suffice or conflict. Our checks use relation knowledge, written by hand or learned from training questions labelled with their relation
paths. It selects a compact context of
facts yielding an answer, and derives it; a derivability check tests whether an answer follows, a consistency check
whether facts conflict, and a validation-chosen routing gate decides where the derived answer replaces the LLM's.
Seven open LLMs read the compact context better in exact match than relevance-ranked facts; on ComplexWebQuestions
(CWQ) they match SubgraphRAG's facts at a quarter of the prompt tokens, and stronger LLMs read SubgraphRAG's better.
Where the routing gate passes, ours leads SubgraphRAG's in exact match in all 53 LLM, benchmark and prompt settings. A
precision gate, returning the derived answer only when the check passes with exactly one derived entity, answered 18
and 5 percent of held-out WebQSP and CWQ questions in a pre-specified test, right for 90 and 96 percent. The
consistency check judges facts by path evidence only where another path joins their entities, true of 2.5 percent of
named-entity facts in our compact WebQSP and CWQ contexts.
\end{abstract}

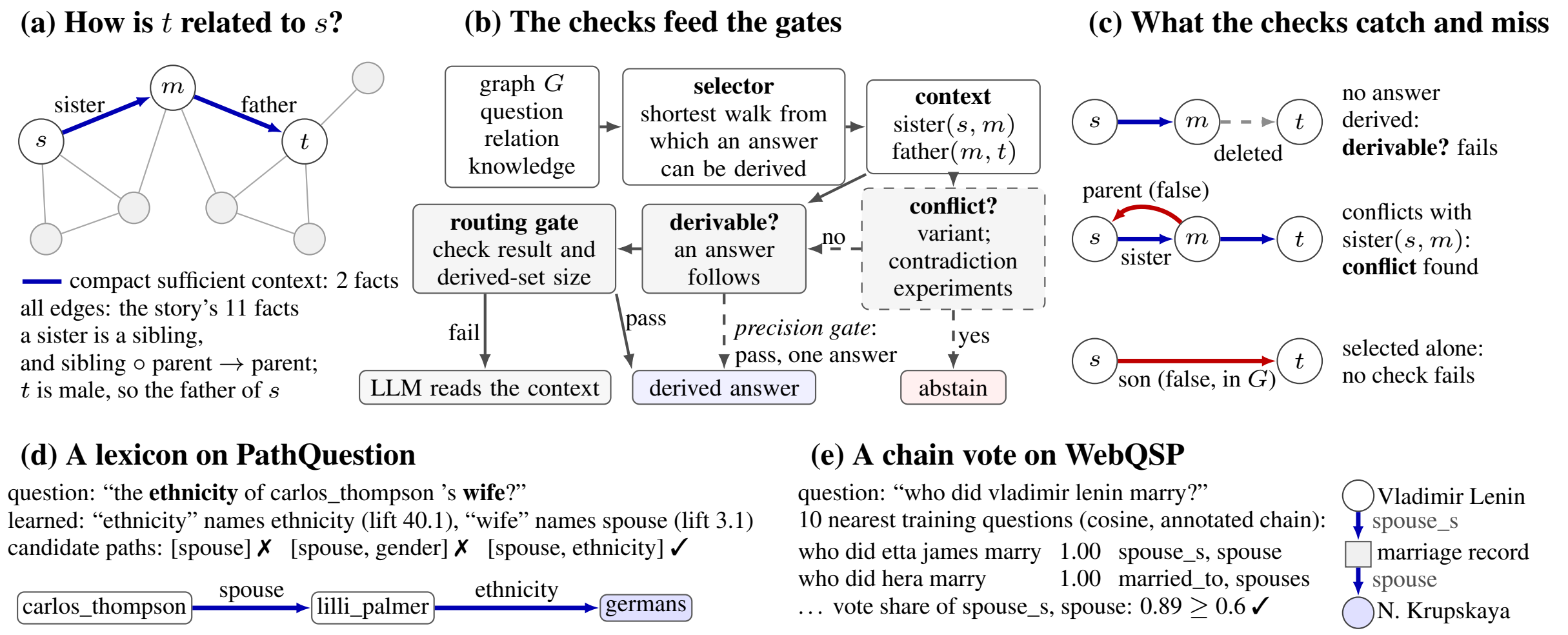
\begin{figure*}[t]
\centering
\begin{tikzpicture}[
  font=\scriptsize,
  ent/.style={circle, draw=black!75, fill=white, minimum size=4.6mm, inner sep=0pt},
  oth/.style={circle, draw=black!40, fill=black!6, minimum size=3.2mm, inner sep=0pt},
  sel/.style={-{Latex[length=1.5mm]}, line width=1.3pt, draw=blue!70!black},
  bg/.style={draw=black!35},
  gone/.style={-{Latex[length=1.5mm]}, line width=1pt, draw=black!45, dashed},
  bad/.style={-{Latex[length=1.5mm]}, line width=1.3pt, draw=red!75!black},
  lab/.style={font=\scriptsize, inner sep=1pt},
  box/.style={draw=black!65, rounded corners=2pt, align=center, inner sep=2.5pt},
  chk/.style={draw=black!65, rounded corners=2pt, align=center, inner sep=2.5pt, fill=black!4},
  ar/.style={-{Latex[length=1.6mm]}, thick, draw=black!70},
  pg/.style={-{Latex[length=1.6mm]}, thick, draw=black!70, dashed},
  head/.style={font=\footnotesize\bfseries, anchor=west},
]
\node[head] at (-0.35,2.5) {(a) How is $t$ related to $s$?};
\node[ent] (s) at (0,1.3) {$s$};
\node[ent] (m) at (1.35,1.85) {$m$};
\node[ent] (t) at (2.7,1.3) {$t$};
\node[oth] (a) at (0.05,0.3) {};
\node[oth] (b) at (0.95,0.62) {};
\node[oth] (c) at (1.85,0.62) {};
\node[oth] (d) at (2.75,0.3) {};
\node[oth] (e) at (3.35,1.95) {};
\draw[bg] (s)--(a); \draw[bg] (s)--(b); \draw[bg] (a)--(b); \draw[bg] (m)--(b);
\draw[bg] (m)--(c); \draw[bg] (t)--(c); \draw[bg] (t)--(d); \draw[bg] (c)--(d); \draw[bg] (t)--(e);
\draw[sel] (s) -- node[lab, above left=-1pt] {sister} (m);
\draw[sel] (m) -- node[lab, above right=-1pt] {father} (t);
\node[align=left, anchor=north west] at (-0.35,0.1)
  {\tikz[baseline=-0.6ex]\draw[line width=1.3pt, blue!70!black] (0,0) -- (0.4,0);
   compact sufficient context: 2 facts\\
   all edges: the story's 11 facts\\
   a sister is a sibling,\\
   and sibling $\circ$ parent $\rightarrow$ parent;\\
   $t$ is male, so the father of $s$};

\node[head] at (4.2,2.5) {(b) The checks feed the gates};
\node[box, text width=14mm] (in) at (4.93,1.45) {graph $G$\\ question\\ relation\\ knowledge};
\node[box, text width=21mm] (selr) at (7.1,1.45) {\textbf{selector}\\ shortest walk from\\ which an answer\\ can be derived};
\node[box, text width=16mm] (ctx) at (9.35,1.45) {\textbf{context}\\ sister$(s,m)$\\ father$(m,t)$};
\node[chk, dashed, text width=17mm] (con) at (9.35,0.2) {\textbf{conflict?} variant;\\ contradiction\\ experiments};
\node[chk, text width=15mm] (suf) at (7.0,0.2) {\textbf{derivable?}\\ an answer\\ follows};
\node[chk, text width=19mm] (gate) at (4.85,0.2) {\textbf{routing gate}\\ check result and\\ derived-set size};
\node[box, text width=24mm, fill=black!3] (llm) at (4.55,-1.22) {LLM reads the context};
\node[box, text width=17mm, fill=blue!6] (rd) at (7.0,-1.22) {derived answer};
\node[box, text width=9mm, fill=red!6] (ab) at (9.35,-1.22) {abstain};
\draw[ar] (in) -- (selr);
\draw[ar] (selr) -- (ctx);
\draw[ar] (ctx.south west) -- (suf.north east);
\draw[pg] (ctx) -- (con);
\draw[pg] (con) -- node[lab, above] {no} (suf);
\draw[ar] (suf) -- (gate);
\draw[pg] (con) -- node[lab, right] {yes} (ab);
\draw[ar] (gate.south -| llm.north) -- node[lab, left] {fail} (llm);
\draw[ar] (gate.south east) -- node[lab, above right=-1pt] {pass} (rd.north west);
\draw[pg] (suf) -- (rd);
\node[lab, anchor=north west, align=left] at ([xshift=2pt, yshift=-1.5pt]suf.south |- con.south) {\emph{precision gate}:\\ pass, one answer};

\node[head] at (10.6,2.5) {(c) What the checks catch and miss};
\node[ent] (s1) at (10.8,1.5) {$s$};
\node[ent] (m1) at (11.85,1.5) {$m$};
\node[ent] (t1) at (12.9,1.5) {$t$};
\draw[sel] (s1) -- (m1);
\draw[gone] (m1) -- node[lab, below=5pt] {deleted} (t1);
\node[align=left, anchor=west] at (13.2,1.5) {no answer\\ derived:\\ \textbf{derivable?} fails};
\node[ent] (s2) at (10.8,0.3) {$s$};
\node[ent] (m2) at (11.85,0.3) {$m$};
\node[ent] (t2) at (12.9,0.3) {$t$};
\draw[sel] (s2) -- node[lab, below=1pt] {sister} (m2);
\draw[sel] (m2) -- (t2);
\draw[bad] (m2) to[bend right=45] node[lab, above] {parent (false)} (s2);
\node[align=left, anchor=west] at (13.2,0.3) {conflicts with\\ sister$(s,m)$:\\ \textbf{conflict} found};
\node[ent] (s3) at (10.8,-0.95) {$s$};
\node[ent] (t3) at (12.9,-0.95) {$t$};
\draw[bad] (s3) -- node[lab, below=1pt] {son (false, in $G$)} (t3);
\node[align=left, anchor=west] at (13.2,-0.95) {selected alone:\\ no check fails};

\node[head] at (-0.35,-1.93) {(d) A lexicon on PathQuestion};
\node[align=left, anchor=north west, inner sep=0pt] at (-0.35,-2.20)
  {question: ``the \textbf{ethnicity} of carlos\_thompson 's \textbf{wife}?''\\
   learned: ``ethnicity'' names ethnicity (lift 40.1), ``wife'' names spouse (lift 3.1)\\
   candidate paths: [spouse] \ding{55} \; [spouse, gender] \ding{55} \; [spouse, ethnicity] \ding{51}};
\node[box, inner sep=1.5pt] (ct) at (0.65,-3.48) {carlos\_thompson};
\node[box, inner sep=1.5pt] (lp) at (3.4,-3.48) {lilli\_palmer};
\node[box, inner sep=1.5pt, fill=blue!12] (ge) at (6.2,-3.48) {germans};
\draw[sel] (ct) -- node[lab, above] {spouse} (lp);
\draw[sel] (lp) -- node[lab, above] {ethnicity} (ge);

\node[head] at (7.75,-1.93) {(e) A chain vote on WebQSP};
\node[align=left, anchor=north west, inner sep=0pt] at (7.75,-2.20)
  {question: ``who did vladimir lenin marry?''\\
   10 nearest training questions (cosine, annotated chain):\\[1pt]
   \begin{tabular}{@{}l@{\hspace{5pt}}r@{\hspace{5pt}}l@{}}
   who did etta james marry & 1.00 & spouse\_s, spouse\\
   who did hera marry & 1.00 & married\_to, spouses\\
   \multicolumn{3}{@{}l}{\dots\ vote share of spouse\_s, spouse: 0.89 $\geq$ 0.6 \ding{51}}
   \end{tabular}};
\node[ent, minimum size=3.2mm] (vl) at (13.5,-2.33) {};
\node[anchor=west, lab] at (13.65,-2.33) {Vladimir Lenin};
\node[draw=black!65, fill=black!6, minimum size=2.6mm, inner sep=0pt] (mr) at (13.5,-2.93) {};
\node[anchor=west, lab] at (13.65,-2.93) {marriage record};
\node[ent, minimum size=3.2mm, fill=blue!12] (nk) at (13.5,-3.53) {};
\node[anchor=west, lab] at (13.65,-3.53) {N.\ Krupskaya};
\draw[sel] (vl) -- (mr);
\draw[sel] (mr) -- (nk);
\node[lab, anchor=west, text=black!70] at (13.6,-2.63) {spouse\_s};
\node[lab, anchor=west, text=black!70] at (13.6,-3.23) {spouse};
\end{tikzpicture}
\caption{Overview. (a) Two of eleven facts form a compact sufficient context, from which the rules derive the answer.
(b) The check's result and the derived set's size feed the routing gate, chosen on validation questions, which returns the derived answer or the LLM's; the precision gate, a separate policy, returns the derived answer only where the check passes with one answer; the consistency check is a variant used in the contradiction experiments, where a flagged context ends in an abstention.
(c) A deleted fact leaves no answer to derive, so the derivability check catches it. The false fact that $s$ is a
parent of $m$ is caught by the consistency check because a second path, the true fact sister$(s,m)$, also joins $s$ and
$m$ and contradicts it; few selected facts have a second path (Section~\ref{sec:checks}). A wrong fact already in the graph,
such as a needed fact whose object we replaced before selection, almost always escapes both checks once selected. (d) The lift of a word for a relation is how many times as often the
relation lies on the annotated paths of training questions containing the word as on those of all training questions:
ethnicity lies on all 15 paths of questions with ``ethnicity'' and on 2.5 percent of all paths, a lift of 40.1. The path
whose relations are exactly those the words name passes (\ding{55}: they differ); its shaded end is the derived answer. (e) Each of the 10
training questions most similar to the question votes for its annotated chain with its cosine similarity as weight;
spouse\_s, spouse gets 89 percent of the total weight (its vote share), above the 60 percent threshold, and is followed.}
\label{fig:concept}
\end{figure*}

\section{Introduction}
\label{sec:intro}

\looseness=-1 Large language models (LLMs) often answer questions from facts that a retriever selects from a knowledge graph
\citep{baek2023,he2024,edge2024,li2025}. Such an answer deserves trust only when the facts are \emph{sufficient}, so
that combining them yields the answer, and \emph{consistent}, so that no fact contradicts another. Relevance ranking
alone establishes neither. Retrievers keep the most relevant facts \citep{baek2023,li2025}, and path generators keep
the paths a fine-tuned model plans \citep{luo2024}. Explorers let the LLM judge when its facts suffice \citep{sun2024},
but LLMs often answer even from contexts that lack a necessary fact \citep{joren2025}.

\looseness=-1 We study what checks outside the LLM establish about both properties and where they fail. The checks rest on
explicit \emph{relation knowledge}, knowledge of what relations mean and how they combine. It takes the form of rules,
such as ``the father of one's sibling is one's own father'', or of a mapping learned from training questions. Such a
mapping learns, for instance, that ``the ethnicity of X's wife'' asks for the relations spouse and ethnicity. A set of facts
is sufficient when this knowledge derives an answer from it. In Figure~\ref{fig:concept}a, two of the eleven facts
suffice to tell how $t$ is related to $s$; they form a \emph{compact sufficient context}. A \emph{selector} searches the
graph for such a context, and the answer the knowledge derives from it is the \emph{derived answer}. A
\emph{derivability check} tests whether an answer follows from a context. A \emph{consistency check} tests whether the
context's facts conflict. \emph{Gates} are decision rules fixed on validation questions. From the check results and the
number of derived answers, a gate returns the derived answer, the LLM's answer or an abstention
(Figure~\ref{fig:concept}b). No LLM is trained.\footnote{Code and data:
\url{https://anonymous.4open.science/r/MSOC-release-3655/}} Without rules, the consistency check is learned from counts
of the relations the graph stores between the two ends of a path; no contradicted example is used to fit or calibrate
it (Section~\ref{sec:method}, Appendix~\ref{app:learned}).

\looseness=-1 Five benchmarks differ in their relation knowledge. CLUTRR \citep{sinha2019} comes with kinship rules.
PathQuestion \citep{zhou2018} and MetaQA \citep{zhang2018} have templated questions. WebQSP \citep{yih2016} and
ComplexWebQuestions (CWQ; \citealp{talmor2018}) have questions people wrote. We test seven open LLMs that answer without
a reasoning phase, as well as GPT-5 \citep{gpt5} and LLMs that reason before answering. Section~\ref{sec:inputs} asks
whether a compact sufficient context helps LLMs read facts. Section~\ref{sec:checks} asks what the checks catch and
miss, and Section~\ref{sec:who} asks who should answer.

\looseness=-1 Answering by following in the graph the relation chain of similar training questions, that is, executing it
as a query, and falling back on an LLM is not new (CBR-KBQA, \citealp{das2021cbr}; DecAF, \citealp{yu2023decaf};
Section~\ref{sec:related}). What is missing is evidence on when each part helps; we contribute three findings.
(1)~We show that facts selected by relation knowledge are easier for LLMs to read: on PathQuestion, MetaQA and WebQSP,
the seven open LLMs answer more accurately in exact match from our compact contexts than from relevance-ranked facts,
with about a tenth to a third of the prompt tokens of SubgraphRAG's facts. (2)~We show that a passing derivability check
is a precision signal, not a guarantee, and turn it into a \emph{precision gate} that answers without an LLM. On WebQSP
and CWQ questions set aside in advance, 81.5 and 62.1 percent of the derived answers that pass the check are right;
also requiring exactly one derived answer raised this to 89.8 and 95.9 percent, on 17.6 and 4.9 percent of the
questions (Tables~\ref{tab:precgate} and~\ref{tab:gates}). (3)~We identify when a consistency check outside the LLM can
work: it judges a fact by path evidence only where a second path in the context joins the fact's two entities. It
detects 83 to 100 percent of the contradictions we inject, most of which have such a path by construction, and we find
that only 2.5 percent of the fact mentions between named entities in our compact WebQSP and CWQ contexts have one,
whereas larger contexts often do.

\section{Related work}
\label{sec:related}

\paragraph{Selecting graph facts.}
\looseness=-1 Most pipelines rank facts by relevance and keep as many as a budget allows. They rank by similarity to the
question \citep{baek2023,yang2024}, gather facts into a connected subgraph or a community summary
\citep{he2024,edge2024}, or score triples with a model or with attention \citep{li2025,xu2026query}. Explorers let the LLM judge after each step whether its facts
suffice \citep{sun2024,chen2024pog}, TRACE \citep{wang2026trace} steers relation choice by earlier explorations, and
Self-RAG \citep{asai2024} trains a critic of its retrieval. We keep the first setting, an LLM reading selected facts,
add the two checks outside the LLM and count how often LLMs answer without a needed fact (Appendices~\ref{app:gating}
and~\ref{app:layers}).

\paragraph{Relation paths and semantic parsing.}
\looseness=-1 Relation paths are generated by a fine-tuned model \citep{luo2024}, retrieved per sub-question \citep{kim2023} or
scored against the question \citep{liu2024path,zhang2024}. WebQSP parsers
execute a parsed query \citep{yih2016}, rank generated logical forms \citep{ye2022rng} or adapt those of similar
questions \citep{das2021cbr}, as our chains do; DecAF \citep{yu2023decaf} combines executed queries with generated
answers, as our routing does. SELF-KBQA \citep{tian2026subgraph} generates a query from retrieved subgraphs and, when
it fails to execute, runs the query of the top-ranked subgraph. We share the executed query and the fallback and add two findings: a query can execute and still misread the question, and on
human-written questions the size of the derived answer set can beat the check as a confidence signal.

\paragraph{Error detection and context effects.}
\looseness=-1 Knowledge graph refinement \citep{paulheim2017} judges single facts against a whole graph by embeddings, local
or path evidence or implying paths \citep{bordes2013,xie2018,melo2020patybred,lao2010,das2017,galarraga2013}. We instead
test a fact against the rest of a context and calibrate the test on consistent contexts, as conformal anomaly detection
does \citep{laxhammar2015conformal,bates2023conformal}. We find that calibrating false alarms to 1 percent over all
contexts leaves them far above 1 percent in the few contexts where the check has evidence to judge a fact. Irrelevant
facts, needed facts in a bad position and the order of premises lower accuracy
\citep{shi2023,wu2024,yoran2024,cuconasu2024,liu2024lost,chen2024premise,saparov2023}; we measure these effects on
contexts known to be sufficient. Knowledge conflicts usually pit the context against the model's training data
\citep{longpre2021,xie2024}; ours lie inside the context. \citet{joren2025} label context
sufficiency with an LLM-based evaluator and analyse how answering models behave under those labels. We instead test
whether a predicted relation query executes on the context, much as justification finding derives the axioms that
entail a consequence \citep{schlobach2003,kalyanpur2007,horridge2008,baader2010,penaloza2017}. Such a query can execute
even when it misreads the question, which Section~\ref{sec:checks} measures.

\section{Method}
\label{sec:method}

\paragraph{Setting and relation knowledge.}
\looseness=-1 A knowledge graph $G$ is a set of triples $(s,p,o)$, each stating that $o$ is the $p$ of $s$. A question asks for
the relation between two entities, $(s,?,t)$, as in CLUTRR, or for the entities reached from a topic entity by a
sequence of relations; a \emph{context} is a subset of $G$ written into the prompt. \emph{Rules} form an ontology: a
hierarchy mapping surface relations to base relations, such as sister to sibling; inverse and symmetry axioms; and
composition rules $p \circ q \rightarrow r$, a $p$ step then a $q$ step being an $r$ step; CLUTRR has such rules.

\paragraph{Learned mappings.}
\looseness=-1 The other benchmarks have no rules, so their relation knowledge is learned from training questions and the
relation paths annotated for them. A \emph{learned mapping} outputs, for a question, the relations it asks for and,
on WebQSP and CWQ, also answer constraints, such as male for ``brother''. The selector then keeps the paths from the
question's topic entity that realise these relations, and their ends form the derived answer. For ``who did vladimir
lenin marry?'', the predicted relations spouse\_s, from a person to a marriage record, and spouse, from the record to each partner,
lead from Vladimir Lenin to Nadezhda Krupskaya, the derived answer (Figure~\ref{fig:concept}e). No step reads a test
question's answer or annotation. On PathQuestion and MetaQA, whose questions follow templates, the mapping is a
\emph{lexicon} that links question words to relations. The \emph{lift} of a word for a relation is how many times as
often the relation lies on the annotated paths of training questions containing the word as on those of all training
questions. A word is linked to a relation when the lift is at least 1.3, a threshold chosen on held-out training
questions (Appendix~\ref{app:setup}). On
PathQuestion's two-hop questions, ethnicity lies on the paths of all 15 training questions containing
``ethnicity'' but of 2.5 percent of all training questions, a lift of 40.1 (Figure~\ref{fig:concept}d). Each word names the linked relation that lies most often on the paths of the training questions containing it. A path passes when its relations are exactly
those the question's words name (Figure~\ref{fig:concept}d). MetaQA also gets a \emph{classifier}, a naive Bayes model
over words and word pairs that predicts the ordered relation sequence. On WebQSP and CWQ, whose questions people wrote,
each training question has an \emph{annotated chain}, the relations and constraints of its annotated SPARQL query. For a
test question we find the 10 most similar training questions by MPNet sentence embeddings \citep{song2020}, with the
topic's name removed. Each of them votes for its own chain with a weight equal to its similarity
(Figure~\ref{fig:concept}e). The winning chain must hold 60 percent of the votes, and a constraint is applied when it
carries at least half of that chain's votes. CWQ uses 20
neighbours, 80 percent and a constraint share of 30 percent, all chosen on validation questions (Appendices~\ref{app:webqsp}
and~\ref{app:cwq}).

\paragraph{Compact sufficient contexts.}
\looseness=-1 A context $C$ is \emph{sufficient} for a question when the relation knowledge derives an answer from $C$ alone; with
several answers, $C$ must yield all that the whole graph yields. The selector enumerates paths from the question's
entities, shortest first: walks of up to ten edges between $s$ and $t$, or paths of up to three edges from a topic
entity. With rules, the selector tests the paths in turn: it maps each relation to its base relation, inverts it where
the path runs against a stored fact, and asks whether the composition rules reduce the sequence to one relation, a parse
that the CYK algorithm \citep{younger1967} performs with the rules as a grammar; it returns the facts of the first path
that reduces, in path order. With a learned mapping the selector returns every path realising the predicted relation sequence
(Appendix~\ref{app:setup}). We call the result a \emph{compact sufficient context}; if no
path passes, the selector returns the best candidate, if any, marked insufficient. The \emph{derivability check} applies
the same derivation to any context, and its result is the \emph{derived answer}.  On WebQSP and CWQ the derived
answer is executed on the full question graph, whereas the context shown to the LLM is cut at 100 facts; the cut can
remove evidence needed to recover the complete answer set (Appendix~\ref{app:executed}).

\paragraph{What a passing check means.}
\looseness=-1 With rules, a passing derivability check means that the rules entail the answer from the context. With a
learned mapping three claims come apart: the predicted relations yield an answer from the context; the context holds
their full answer set, as it does when the selector built it and nothing was cut; and they are the relations the
question asks for, which can fail. The consistency check, a
separate test, can flag a passing context (Appendix~\ref{app:layers}). On questions not used for testing, the
derived answer where the check passes is right for over 95 percent with CLUTRR's rules, MetaQA's classifier and
PathQuestion's lexicon, but for 85.0 and 61.6 with WebQSP's and CWQ's chains (Appendices~\ref{app:setup}
and~\ref{app:cwq}). The PathQuestion, WebQSP and CWQ questions among them also set the lexicon's lift and the chains'
settings. 

\paragraph{Consistency checks.}
\looseness=-1 A consistency check asks whether a fact of the context contradicts the others. With rules, it composes, for
every fact $(s,p,o)$, the other facts into a relation between $s$ and $o$ and detects a conflict when that relation
disagrees with $p$. In Figure~\ref{fig:concept}c, a false fact states that $s$ is a parent of $m$, whereas the rules
derive from the fact that $m$ is the sister of $s$ that the two are siblings. Without rules, a learned check estimates
from the graph which relations hold between the two ends of a path. Random walks over the graph count, for each
sequence of relations, which relation the graph stores between the walk's two ends; a fact $(s,p,o)$ is suspicious when
$p$ is rarely stored at the ends of the paths that the other facts form between $s$ and $o$. Appendix~\ref{app:learned} works through an
example. A second score asks the same of TransE relation vectors \citep{bordes2013}, under which a path's
relations add up to the relation it implies.

\looseness=-1 A context takes the score of its most suspicious fact, as one
false fact suffices, and this score is compared with consistent calibration contexts of similar size: the context's
\emph{share} is the fraction of them that score at least as suspiciously, so a small share marks an unusual context.
The context is flagged when the smaller of its two shares falls below a threshold set so that 1 percent of all
calibration contexts are flagged. Separately, a relation that has one value for at least 99 percent of its subjects in
the graph, such as a person's place of birth, is treated as single-valued, and a context that gives a subject two
values of it is flagged. No contradicted example is used to fit the scores or to calibrate the threshold; the decision
rule was chosen on synthetic contradictions in training contexts and the default configuration post hoc on the probe
results of Appendix~\ref{app:learned}.

\paragraph{Deciding who answers.}
\looseness=-1 \emph{Routing} returns the derived answer where the
\emph{routing gate} passes and the LLM's answer elsewhere; it reads the derivability check and the derived set only. The
consistency check enters as a variant (Table~\ref{tab:decomp}) and in the contradiction experiments, where a flagged
context ends in an abstention reporting the conflicting facts (Table~\ref{tab:faith}). On PathQuestion and MetaQA the routing gate is the derivability check. On WebQSP
and CWQ we choose it among 43 candidate rules, each deciding from whether the check passes and how many answers are
derived when to return the derived answer: ``the check passes'', ``at most $t$ answers are derived'', ``the check passes
and at most $t$ are'' and ``the check passes or at most $t$ are'', with $t$ from 1 to 10, 15, 20, 50 or unlimited. On
validation questions we take the rule under which routing gives the seven open LLMs the highest mean exact match. It is ``the check passes or there are at most 20 derived
answers'' on WebQSP and ``exactly one derived answer'' on CWQ, the \emph{one-answer gate} (Table~\ref{tab:gates},
Appendix~\ref{app:routing}).
The \emph{precision gate} requires the check to pass and one derived answer; it is meant for answering without an LLM
where precision matters more than coverage. The 43 rules were defined after test comparisons and the precision gate
after its CWQ test precision had been seen; held-out questions then confirmed both gates' pre-set precision criteria
(Section~\ref{sec:checks}).

\section{Experimental setup}
\label{sec:setup}

\tabinputs

\paragraph{Benchmarks.}
\looseness=-1 CLUTRR generates family stories and asks how two people are related, with 22 kinship labels; we use 4,361
test questions with chains of 2 to 8 facts. PathQuestion asks two- and three-hop questions about people in Freebase
\citep{bollacker2008}, a large collaborative knowledge graph (782 test questions), and MetaQA one- to three-hop questions
about films in WikiMovies \citep{miller2016}, a film knowledge graph (900). WebQSP has 1,628 test questions written from
search queries, and CWQ 3,531 that extend WebQSP's queries by composition, conjunction, comparison and superlatives; the
CWQ questions built from one WebQSP question form a \emph{seed cluster}. Both use the Freebase question graphs released
with RoG.

\paragraph{Contexts and baselines.}
\looseness=-1 We compare our context with all facts of the story on CLUTRR or the one-hop neighbourhood of the question's entities,
cut at 300 facts on MetaQA and WebQSP; and the $k$ facts most similar to the question under Sentence-BERT sentence
embeddings \citep{reimers2019}, with $k$ our context's size. We also use the annotated path, a reference context built
from the question's gold relation path, and four published methods (Appendix~\ref{app:baselines}): KAPING \citep{baek2023}, up to 10 facts; SubgraphRAG \citep{li2025}, up to 100, its
retriever retrained on PathQuestion and MetaQA, also cut to our context's size and to smaller budgets;
Think-on-Graph \citep{sun2024}, at width and depth 3; and RoG \citep{luo2024}, which plans relation paths. Every context
goes into the same prompt for the same LLMs; on WebQSP ours and SubgraphRAG's facts are also read with its own prompt.

\paragraph{Probes.}
\looseness=-1 Probes alter compact contexts to test what the checks catch and what LLMs do without them
(Appendix~\ref{app:setup}). On CLUTRR we delete a necessary fact, giving a \emph{depleted context}; replace one
relation by friend, neighbour or coworker; add 2 to 16 distractors; shuffle the facts; or add a false fact: a
\emph{reversed-edge contradiction} adds, to the fact that $m$ is the sister of $s$, that $s$ is a parent of $m$, and a
\emph{far contradiction} relates two people at least two steps apart against the rules. PathQuestion and WebQSP get their own
probes (Appendix~\ref{app:setup}), and on all three false facts are also put in the graph before selection.

\paragraph{Prompts and scores.}
\looseness=-1 The \emph{forced-choice prompt} requires an answer; the \emph{abstention prompt} adds the reply that the
facts are insufficient, and the \emph{status prompt} also the reply that they contradict each other
(Appendix~\ref{app:setup}). Answer sets are scored by exact match, Hit (some gold answer) and F1; answers of GPT-5 and Qwen3-32B cut
at the output limit in Tables~\ref{app:passfail:split} and~\ref{app:reason:full}, re-asked with a larger limit, move no
difference by over 1.3 points (Appendix~\ref{app:caps}).

\paragraph{Models and statistics.}
\looseness=-1 We use seven open instruction-tuned LLMs, Llama-3.2-3B \citep{llama32}, Llama-3.1-8B \citep{grattafiori2024}, Qwen2.5-7B
\citep{qwen25}, EXAONE-3.5-7.8B and EXAONE-3.5-32B \citep{exaone35}, Gemma-3-27B \citep{gemma3} and Qwen3-32B
\citep{qwen3}, quantised to 4 bits and decoded greedily without reasoning. On a 564-question subset (264 CLUTRR, 150 PathQuestion and 150 MetaQA) we add Qwen3-32B with reasoning,
DeepSeek-R1-Distill-Llama-70B \citep{deepseekr1}, GPT-5 with and without reasoning, and GPT-5-mini
\citep{gpt5}; GPT-5 also reads every WebQSP and CWQ question, with and without reasoning, and GPT-5 and Qwen3-32B
with reasoning every PathQuestion, MetaQA and CWQ question (Appendices~\ref{app:cwq} and~\ref{app:reason}).
Paired tests are Holm-adjusted over the LLMs compared, and brackets give 95 percent confidence intervals
(Appendix~\ref{app:setup}).

\section{Result 1: when compact contexts help}
\label{sec:inputs}

This section asks whether the same LLM under the same prompt answers more accurately from the compact context than
from the facts other methods select; we compare exact match per LLM and its mean over the seven open LLMs
(Table~\ref{tab:inputs}).

\paragraph{Questions written by people.}
\looseness=-1 On WebQSP, our contexts follow the chains that our learned mapping predicts and hold 14.7 facts on average. The
seven open LLMs read them more accurately in exact match than SubgraphRAG's 100 facts, by 10.2 points on average
(Table~\ref{tab:inputs}) and by 4.6 to 13.0 points for each LLM, and than RoG's planned paths; SubgraphRAG's facts hold
every gold answer more often and give more hits (Appendices~\ref{app:baselines} and~\ref{app:webqsp}). On CWQ, 18.6-fact contexts are read by the seven open LLMs about as accurately in exact match as
SubgraphRAG's 100 facts at a quarter of the prompt tokens, with lower Hit and F1, and above RoG's paths. Strong LLMs
read SubgraphRAG's facts better, by 3.5 to 7.6 points of exact match (Appendix~\ref{app:cwq}); on all WebQSP questions GPT-5 with reasoning reads both alike in exact
match (Appendix~\ref{app:reason}). Most of the gain needs annotated chains: with chains from shortest
question-to-answer paths instead (weak supervision) the open LLMs lead SubgraphRAG's facts on WebQSP only in exact match (47.4
against 43.9) and trail on CWQ in all three metrics
(Appendix~\ref{app:passfail}). With our context, SubgraphRAG's own prompt lowers every open LLM's exact match by 3.5
to 15.7 points against our forced-choice prompt (Appendix~\ref{app:baselines}).

\paragraph{Where the routing gate passes, the compact context leads.}
\looseness=-1 SubgraphRAG's Hit lead comes mostly from questions that fail the routing gate (Appendix~\ref{app:passfail}).
Where it passes, our context leads SubgraphRAG's 100 facts in exact match in all 53
settings, each one LLM reading one benchmark under one prompt, significantly in 52 of the 53, by a median of 15.1 points
(1.2 to 34.4), and in F1 in 50 of the 53, trailing significantly in 2 (Table~\ref{app:passfail:split}, whose caption also gives the
pass rates; Appendix~\ref{app:passregion}). Settings share LLMs, so we also count units: all 9 LLMs lead, as do all 6 model series (sign
tests, $p \leq 0.031$). On WebQSP, MetaQA and CWQ the
passing questions are easier for SubgraphRAG too, so the lead is conditional, not causal; its facts cut to our context's
size still lose there, by 19.8 to 25.0 points for every open LLM on WebQSP (Appendix~\ref{app:passregion}).

\paragraph{What each part contributes.}
\looseness=-1 On WebQSP our context instead of SubgraphRAG's facts gains the open LLMs 10.2 points of mean exact match,
and routing to the derived answer adds 5.0; the gain thus comes from selecting the context and from returning the
derived answer; picking the questions that get the derived answer by the vote share instead of the routing gate
changes it by under one point (Table~\ref{tab:decomp},
Appendix~\ref{app:passfail}).

\paragraph{Templated and rule-made questions.}
\looseness=-1 In exact match the compact context beats KAPING's, Think-on-Graph's and SubgraphRAG's facts for every LLM
on PathQuestion and MetaQA, by about 7 points or more ($p \leq 7\times10^{-6}$), and GPT-5 and Qwen3-32B with reasoning
read it better than SubgraphRAG's facts on all their questions, by 5.0 to 12.9 points
($p \leq 7.5\times10^{-6}$), with fewer hits on MetaQA (Appendix~\ref{app:reason}). It needs one LLM call and about a tenth to
a third of SubgraphRAG's prompt tokens (Table~\ref{tab:inputs}).
On CLUTRR the compact context, which for 96.6 percent of questions is the shortest path between the two people, beats
the whole story by 2.1 to 5.2 points for six LLMs; there the rules matter for the checks, not for finding the facts
(Appendices~\ref{app:clutrr} and~\ref{app:setup}).

\section{Result 2: what the checks catch}
\label{sec:checks}

This section asks how often a passing derivability check has a right derived answer per kind of
relation knowledge, whether a cheaper estimator does as well and what the consistency check catches.

\paragraph{The derivability check is a precision signal.}
On CLUTRR, \emph{gating} an LLM on the check returns its reply where the check passes and abstains otherwise. It raises
the LLM's \emph{balanced score}, the mean of its accuracy on sufficient contexts and its abstention rate on insufficient
ones, by up to 31 and 37 points with a necessary fact deleted or a relation replaced. On the 330
strict two-fact questions, whose kinship label real families cannot change, the three largest LLMs already abstain on
depleted contexts themselves, so gating raises their balanced score by at most 2 points (Appendices~\ref{app:gating}
and~\ref{app:depth}). On WebQSP a pass
shows that the predicted chain can be followed in the context to an answer, not that it is what the question
means: with its constraints, it matches the question's annotated chain for 79.3 percent of passing questions (Appendix~\ref{app:layers}).

\looseness=-1 Is the check the best signal for trusting the derived answer? We compare it with seven other confidence
signals, such as the vote share alone and SubgraphRAG's top-1 triple score, by \emph{risk}, the share of returned
derived answers that are wrong, at equal \emph{coverage}, the share of questions given one. On WebQSP, at the check's
coverage of 45.6 percent, the check has the lowest risk, 21.3 percent against 23.8 for the vote share alone;
SubgraphRAG's score is level with it over the coverage range all signals reach (Appendix~\ref{app:riskcov}). On CWQ the
number of derived answers is the better signal: at the check's coverage of 11.8 percent, returning the questions with
the fewest derived answers has risk 14.1 percent against the check's 48.0. Chosen on validation questions, CWQ's
one-answer gate raises the open LLMs' exact match under routing by 2.9 points [+2.4, +3.5] over routing on the check
(Appendix~\ref{app:routing}).

\tabprecgate

\paragraph{Confirmed on held-out questions.}
The gates were chosen on validation questions, and the precision gate was named after its CWQ test precision had been
seen, so their precision could be overstated. We therefore tested them on \emph{held-out questions}: training questions
never used for evaluation before this test, 500 from WebQSP and 1,501 from CWQ, removed with their
near-duplicates from the pool of training questions that vote for chains (Appendix~\ref{app:prereg}). Before selecting them, we fixed the gates and four
hypotheses in a plan: (H1) each routing gate's precision is at or above the lower end of the 95 percent Wilson interval of
its validation precision, 61.6 and 66.0 percent, and a one-sided test puts it above 50 percent; (H2) routing beats each
open LLM alone; (H3) the routing gate routes better than the check; (H4) the precision gate's precision exceeds 80
percent. H1 held, with 66.0 and 69.0 percent, and so did H4, with 89.8 and 95.9 (Table~\ref{tab:precgate});
Section~\ref{sec:who} reports H2 and H3.

\paragraph{What the consistency check can judge.}
One reversed-edge contradiction costs an open LLM 32 to 42 percent of its accuracy (Appendix~\ref{app:clutrr}), few
open LLMs report the conflict (Appendix~\ref{app:reason}), and the derivability check, which tests only whether an
answer follows, passes 99.9 percent of contradicted contexts; catching the contradiction is the consistency check's job.
The consistency check with rules detects at least 99.8 percent of them with at most 0.8 percent false alarms, lifting
the \emph{three-way score}, the mean of the balanced score's two rates and the abstention rate on contradicted
contexts, to between 0.73 and 0.90 (Table~\ref{tab:faith}), and the learned check 83 to 100 percent
(Table~\ref{tab:learned}, Appendix~\ref{app:learned}).
These rates hold only where the context holds a \emph{second path}, a chain of other facts joining the two entities of
the fact under test, as sister$(s,m)$ joins $s$ and $m$ in Figure~\ref{fig:concept}c; the injected contradictions other than second
values create one by construction. Our WebQSP and CWQ contexts hold the paths from the topic entity to the answers and a
few facts of the records and constraints on them, which rarely join two entities twice: only 2.5 percent of the fact mentions between named entities have a second path
(Table~\ref{tab:checkable}). Compact contexts are sparse by design: some fact has a second path in 2.6, 3.0 and 26.9
percent of the WebQSP, CLUTRR and PathQuestion compact contexts, but in 92.7 percent of SubgraphRAG's 100 WebQSP facts
(Appendix~\ref{app:checkable}). The single-value rule needs no second path; it applies to another 2.2 percent of mentions,
whose relations are single-valued, and finds no second value among them. Where a second path exists, the check flags
27.6 percent of the mentions not confirmed as errors and 29.9 percent of the contexts. The 1 percent false-alarm rate
holds over all consistent calibration contexts; most of them have no second path, and none of those was flagged, while
27.0 percent of those with a second path are. Around natural errors
that LLMs judged against the knowledge base Wikidata \citep{vrandecic2014}, the check flags errors and supported facts
alike (area under the receiver operating characteristic curve, AUC, 0.52; Appendix~\ref{app:checkable}).
If the graph itself holds a wrong fact, as when we replace the object of a needed fact before selection, and the
selector picks it, both checks almost always miss it (Appendix~\ref{app:learned}).

\section{Result 3: who should answer}
\label{sec:who}

\tabanswer

\paragraph{Derived answers are most accurate on benchmarks with reliable relation mappings.}
Table~\ref{tab:answer} compares, per benchmark, the derived answer with the best open LLM and GPT-5 where the routing
gate passes (routing on all questions: Table~\ref{tab:answerfull}). With CLUTRR's rules and MetaQA's classifier
almost all contexts pass and the derived answer is above every LLM; with PathQuestion's lexicon 77 percent pass, the
derived answer beats every LLM, and routing beats every LLM alone, GPT-5 by 3.3 points (Appendix~\ref{app:reason}). On CLUTRR the derived answer applies the rules that generated the
labels, which the LLMs are not given; given them under the status prompt, GPT-5 with reasoning answers 89.0 rather than
44.7 percent of the subset correctly (Appendix~\ref{app:reason}). With WebQSP's
and CWQ's chains, less accurate behind the check, the derived answers are level with the three largest open LLMs and at most
level with GPT-5 where the routing gates pass. Behind the precision gate they are
right for 91.3 and 96.4 percent (Table~\ref{tab:precgate}), level with or above every LLM there
(Appendices~\ref{app:cwq} and~\ref{app:routing}).

\paragraph{On WebQSP and CWQ, validation questions choose who answers.}
Validation questions choose, per open LLM, one of six policies, such as the LLM alone or routing (all six:
Appendix~\ref{app:riskcov}), by accuracy or by the \emph{equal-cost score}, which counts $+1$ for a right answer, $-1$ for a wrong one and 0 for an abstention; test and
held-out questions then score it by the same criterion. No chosen policy was significantly worse than
the LLM alone on the criterion that chose it. On held-out questions, routing raised the accuracy of the four smaller open LLMs by 4.7 to 9.3 points and
changed that of the three largest by $-0.8$ to $+1.4$, so H2 held only for the smaller
four. H3 held on CWQ (+2.6 [+1.4, +3.9]) but was not distinguished on WebQSP (+0.3
[$-1.5$, +2.1]; Appendix~\ref{app:prereg}). For GPT-5 without reasoning, a near tie on WebQSP validation chose
routing, which cost 2.7 points on test ($p = 6.3\times10^{-4}$).

\section{Conclusion}
\label{sec:concl}

Relation knowledge selects compact contexts that open LLMs read better in exact match than relevance-ranked facts on
PathQuestion, MetaQA and WebQSP, with far fewer prompt tokens than SubgraphRAG's; on WebQSP most of this gain needs
annotated chains, and on CWQ strong LLMs read SubgraphRAG's facts better. A passing derivability check does not guarantee a right answer, but a
precision gate fixed in advance met its precision criterion on a small share of held-out questions, and no
policy chosen per open LLM on validation questions was significantly worse than that LLM alone on its own criterion. A consistency check outside the LLM can judge a fact by path
evidence only where a second path joins its entities, which few facts in our contexts have; graph contamination remains open.

\section*{Limitations}
\label{sec:lim}

\looseness=-1 The reach of the derived answer on human-written questions is small and depends on seen chains. Of
WebQSP's test questions, 94.7 percent have a relation chain seen in training; the 63 with only unseen chains pass the
check for 4 questions, all wrong, and two-relation chains pass with 60.7 percent right against 87.0 for one relation
(Appendix~\ref{app:webqsp:scope}). On CWQ the check passes 11.8 percent and the one-answer gate 24.5 percent, and no
comparative or superlative question passes with a right answer (Appendix~\ref{app:cwq}). Our derived answers reach
64.4 F1 on WebQSP, below semantic parsers such as RnG-KBQA \citep{ye2022rng} and DecAF \citep{yu2023decaf}, which
report 75.6 and 77.1 to 78.8 \citep{yu2023decaf}; SubgraphRAG is the only strong retriever we run, and the others are
cited. The chains come from annotated training questions or SPARQL queries; weakly supervised chains read 6.0 to 7.6
points worse on WebQSP and trail SubgraphRAG's facts there in Hit and F1, and on CWQ in all three metrics.

WebQSP and CWQ use RoG's question graphs, which hold no gold answer for 19 percent of CWQ's test questions and keep no
dates or numbers. Questions whose gold query has a date, numeric, ordering or literal constraint, 10.3 percent of WebQSP
and 28.6 of CWQ, pass with 22.0 and 3.1 percent of derived answers exact, against 84.1 and 61.1 on reachable questions
without one. Literals and unreachable answers explain a minority of the wrong passing answers; the rest, 108 and 137,
are mostly wrong chains, constraints or answer sets, with other graph gaps such as missing or renamed answers
(Appendix~\ref{app:layers}).

The checks read the compact context, so a false fact that becomes the selected path escapes them: a substituted relation
or object in the graph yields an undetected wrong answer for 97.9, 41.1 and 43.1 percent of the CLUTRR, PathQuestion and
WebQSP questions given one (Appendix~\ref{app:learned}). Apart from the natural-error study, missing, false and irrelevant facts are studied through
synthetic probes. The natural-error labels are LLM judgements against Wikidata, not the authors':
two LLM passes, the first by the same LLM that drafted the labelling rules, then a tie-break; an independent GPT-5
pass leaves the conclusion unchanged but changes the error count, so the labels should not be reused as verified facts
(Appendix~\ref{app:learned}). The learned check detects only contradictions the graph expresses through co-stored
relations, paths or single values, is calibrated for small contexts, and its CLUTRR training families come from the
benchmark's generator. CLUTRR's labels come from the rules the checks use, so there the derived answer shows agreement
with the rules rather than real kinship, and about 47 percent of its questions admit a step or in-law reading
(Appendix~\ref{app:depth}).

The open models, up to 32 billion parameters and quantised to 4 bits, cover the full study. Reasoning mode and the
proprietary models cover a 564-question subset; GPT-5 also reads every WebQSP and CWQ question, and GPT-5 and Qwen3-32B
with reasoning every PathQuestion, MetaQA and CWQ question. Reasoning LLMs report contradictions themselves, at about
250 to 720 output tokens per question. The benchmarks are public and may be in the LLMs' training data; on WebQSP the
open LLMs answer rightly from their own knowledge for at most 4.7 percent of depleted contexts
(Appendix~\ref{app:layers}). The study is in English, the prompts were not tuned per model, repeated runs vary by up to
1.2 points, and the selector presumes that the question's entities are identified. Apart from the post hoc 90 percent
line, the choice of path counts, the gates' candidate family and the precision gate, defined after test comparisons,
the constants were set by hand on training or validation data; Appendix~\ref{app:setup} gives their sensitivity and the
steps for a new graph.

\section*{Ethics statement}
\label{sec:eth}

All data are synthetic or come from public benchmarks; the knowledge-base facts about real, mostly public, people are
used only as released, and no new personal data is collected. The checks let a pipeline abstain rather than guess, but
they are only as sound as the relation knowledge behind them, and trusting them beyond that is a risk: incomplete
knowledge reports some sufficient contexts as insufficient, wrong knowledge returns a wrong answer as derived, and a
learned consistency check misses contradictions of kinds that its graph does not express. A passing check should be
shown to users as an answer derived under the predicted relations, with no conflict of the supported kinds found, not
as a verified answer.

\paragraph{Use of AI assistants.}
We used generative AI tools to help write code, draft and revise the manuscript, and find and correct errors. The
authors reviewed all code, results and text and take full responsibility for the content of this paper.

\bibliography{msoc}

\appendix

\providecommand{\supp}[1]{}

\section{Setup details}
\label{app:setup}

\paragraph{Complete tables.} \looseness=-1 The appendices state the numbers the paper cites; the supplementary material gives, under the same letters, every per-LLM and per-condition table and the details the appendices defer to it. Beyond Section~\ref{sec:setup}, CLUTRR adds 2, 4, 8 or 16 distractors attached to the path or among people
absent from the story, and 1, 2 or 4 reversed-edge or 1 or 2 far contradictions. PathQuestion reverses a family fact or makes a person their own grandparent's parent; WebQSP reverses a family or containment fact, gives a single-valued relation a second value or deletes a fact of the predicted chain's paths; Appendix~\ref{app:learned} enters false facts
before selection (reversed edges and facts, far facts, shortcuts, substitutions, second values). All conditions of a
benchmark are built from the same graph and questions, so every comparison is paired.

\paragraph{Prompts.} \looseness=-1
On CLUTRR the prompt restricts the answer to exactly one of the 22 kinship labels, lists the facts as one sentence each and ends with the question (verbatim: the supplementary material).\supp{On CLUTRR the prompt reads:
{\footnotesize\begin{list}{}{\setlength{\leftmargin}{1em}\setlength{\rightmargin}{0pt}\setlength{\topsep}{3pt}}\item[]\ttfamily\raggedright
You answer family-relation questions using ONLY the FACTS below.\\
Answer with EXACTLY ONE label from: {\rmfamily\itshape the 22 kinship labels}.\\
{[}If the FACTS are insufficient to determine the relation, output exactly: INSUFFICIENT\_CONTEXT{]}\\
Output ONLY the label, nothing else.\\
FACTS\\
- Mavis's father is Tom.\ \ {\rmfamily\itshape one line per fact}\\
QUESTION\\
How is Tom related to Anna? (Complete the sentence: Tom is Anna's \_\_\_.)
\end{list}}}
On PathQuestion, MetaQA and WebQSP the prompt states that each fact is written as \texttt{(subject) -{}-[relation]-{}-> (object)}, asks for the answer entities exactly as they appear in the facts, separated by `` | '' when several are
correct, and ends with the question. The line permitting the reply INSUFFICIENT\_CONTEXT appears only in the abstention prompt, to which the status
prompt adds, before it, the line ``If the FACTS contradict each other, output exactly: CONTRADICTORY\_CONTEXT'';
applied unchanged to clean, insufficient, contradicted and distracted contexts, the status prompt lets an LLM's
reports of a contradiction be counted as detections and as false alarms.

\paragraph{Relation knowledge.} \looseness=-1
The CLUTRR rules are the benchmark's own relation store: 11 base relations, each with a male and a female label, 4 inverse axioms, 2 symmetry axioms (we add a third for sibling-in-law, which the store omits) and 16 composition rules.\supp{The base relations are child, parent, spouse, sibling, grandchild, grandparent, child-in-law, parent-in-law, sibling-in-law, nephew or niece and uncle or aunt; the symmetry axioms cover sibling and spouse.} The rules omit some
compositions, such as the uncle of a sibling, so the shortest path cannot be composed for 158 questions
(3.6 percent); for 149 of them the selector finds another composable path, and for the other 9 it returns
the shortest path marked insufficient, so 4,212 compact contexts (96.6 percent) are the shortest path. About 47 percent of the questions involve one of four compositions, child of
a grandparent, child of an uncle or aunt, spouse of a child and spouse of a grandchild, whose CLUTRR answer admits a
step or in-law reading; Appendix~\ref{app:depth} repeats the main comparisons on questions free of them.
PathQuestion's lexicon is $P(\text{relation family}\mid\text{question word})$ fitted on training questions and their
annotated paths, a relation family grouping Freebase relations ending in the same name\supp{A relation family groups Freebase relations ending in the same name, such as film.film.country and location.administrative\_division.country, which a word like ``country'' cannot tell apart.}; it keeps only words that raise a family's probability to at least 1.3 times its base rate (the lift) and is complemented by overlap of
relation names with question words. MetaQA's lexicon is of the same kind, and its supervised classifier is a naive
Bayes model over question n-grams trained on about 5,000 questions per hop subset\supp{The classifier is trained on 4,719, 4,999 and 5,000 questions of the one-, two- and three-hop subsets (5,000 sampled, less those dropped below), with 13, 21 and 15 sequence classes.}. The hops of a MetaQA question's subset bound the length of the relation sequence;
PathQuestion searches paths of up to three edges. Appendices~\ref{app:webqsp} and~\ref{app:learned}
give WebQSP's relation knowledge and the learned consistency check.

\paragraph{Benchmark details.} \looseness=-1
From PathQuestion we use PQ-2H and PQ-3H, two- and three-hop questions over graphs of 1,211 and 2,839
triples, and PQL-2H, two-hop questions of the larger PathQuestion-Large (PQL) release over 4,247 triples. Questions whose gold path contains a triple
from an entity to itself, an artefact of entity names that collide in the released graphs, are dropped from the test sets and SubgraphRAG's training but kept for fitting the
lexicon, which uses only their words and relations; PQL-3H, where most questions are affected, is dropped.\supp{The artefact affects 6, 4 and 818 of the 1,908, 5,198 and 1,594 questions of PQ-2H, PQ-3H and PQL-2H, and 956 of PQL-3H's 1,031.} Training and test questions
have disjoint topic entities (80/20); the test sets are 300 sampled questions each from PQ-2H and PQ-3H and all 182
remaining test questions of PQL-2H (of 348). \supp{Of the 782 questions, 189 have several gold answers and 40 have 20 answers, which exceed the output limit of 64 tokens and become solvable once capped answers are regenerated with 512 tokens (below).} MetaQA uses the WikiMovies graph of 133,582 triples over 9 relations. Its annotated paths execute each
question's released relation chain on the graph, keeping every branch, and reproduce the released answer set for all but a few one- and two-hop questions, which are dropped.\supp{They reproduce it for every three-hop question, all but one of the 14,872 two-hop test questions and 94.5 percent of the one-hop ones; the others are mostly one-hop questions about films sharing a title.} WebQSP uses the 1,628 test questions and the per-question Freebase
subgraphs released with RoG (2,826 training and 246 validation questions), which SubgraphRAG also uses; the topic
entities are given, and the subgraphs keep no literals such as dates. CWQ uses RoG's graphs too (Appendix~\ref{app:cwq}).

\paragraph{Models and decoding.} \looseness=-1
The open models run on Ollama 0.9.6,\footnote{\url{https://github.com/ollama/ollama}} which serves them through llama.cpp,\footnote{\url{https://github.com/ggml-org/llama.cpp}} in its Q4\_K\_M quantisation, a 4-bit weight format, decoding greedily at
temperature 0 with a fixed seed. Outputs are limited to 24, 64, 256 and 128 tokens on CLUTRR, PathQuestion, MetaQA, and
WebQSP and CWQ (512 for the status prompt on PathQuestion); answers that reached their limit were generated again with 512 tokens (1,024 on MetaQA), which, up to
the decoding noise reported below, equals running every condition with the larger limit, as an answer that ended earlier stays the same, with the exceptions listed in the supplementary material; other runs used the larger limits from the start.\supp{ A few runs kept their limits (54 answers, at most 0.5 percent of a condition).}\supp{For 43 such answers of Think-on-Graph, whose final prompt is built during its search, the prompt was rebuilt from the stored facts and used only if the run's cache held the old answer under it. The status prompt's runs on CLUTRR outside reasoning mode keep 24 tokens, and Qwen3-32B, as the exploring LLM of the supplementary material, keeps 128 for every call; the answers that reached these limits, 52 of 186,330 (four LLMs, at most 0.5 percent of a condition) and 2 of its 300 (none of its steps), were not generated again.} \supp{The larger limit still cuts 0.4 percent of the WebQSP answers, 0.2 of the MetaQA and CWQ ones and almost none of the others, mostly repeating entities already listed.}
The context windows are 4,096 tokens on CLUTRR, 8,192 on MetaQA and 16,384 on PathQuestion, WebQSP and CWQ. \supp{On CWQ, Gemma-3-27B and Qwen3-32B run with 8,192, as does Qwen3-32B as the exploring LLM and on WebQSP's kNN-chain test contexts under the forced-choice prompt (sharing MetaQA's runner), more than any of these prompts needs (at most 3,636 tokens).} Gemma-3-27B runs with 8,192 on
PathQuestion and WebQSP, as its runner failed with longer windows, so it could not read the two-hop neighbourhood on
PathQuestion (cut at 600 facts) or the one-hop one on WebQSP, whose prompts reach 15 and 11 thousand tokens. Byte-identical
prompts in two conditions differ in net accuracy by at most 0.5 points on CLUTRR, 1.2 on PathQuestion and 1.0 on
MetaQA. In reasoning mode, Qwen3-32B runs with thinking enabled and DeepSeek-R1-Distill-Llama-70B reasons inside think
tags removed before the answer is parsed; both may write up to 4,096 tokens in a window of 8,192 (6,144 and 12,288 on MetaQA); at most 14.4 percent of a condition's answers reached that limit\supp{ (265 of 13,308 on the subsets of Appendix~\ref{app:reason} and under the status prompt, 271 of Qwen3-32B's 12,674 on the PathQuestion and MetaQA full sets, 220 of its 7,062 on CWQ, bounded in Appendix~\ref{app:cwq})}, and the limit bounds the reasoning, as GPT-5's does; Appendix~\ref{app:caps} re-asks Qwen3-32B's 271 capped full-set answers with 8,192 tokens.\supp{Of the 265 and 271 answers at the limit, 214 and 264 are empty. Byte-identical prompts shared one cached completion, except with the runs on KAPING's, the similarity top-$k$ and Think-on-Graph's facts, done in parts with a cache each, so the 12,674 full-set answers are 9,182 prompts and at least 9,745 distinct completions (240 at the limit, 235 empty); 543 prompts were called more than once in these runs and 457 have differing completions, 24 on PathQuestion disagreeing on the limit.}
GPT-5 and GPT-5-mini run through the OpenAI interface at reasoning effort minimal with 256 output
tokens, and GPT-5 with reasoning at effort low with 2,048; GPT-5's runs under the status prompt and on WebQSP's validation questions allow 4,096 in both settings.
\supp{GPT-5 reads all 1,628 WebQSP test questions with our context and SubgraphRAG's top 10, 20, 50 and 100 triples in both settings, and with reasoning all 782 PathQuestion and 900 MetaQA questions with Table~\ref{tab:inputs}'s contexts and MetaQA's classifier context, Think-on-Graph's facts being those Qwen3-32B collected answering directly; it reads all 3,531 CWQ test questions with our context and SubgraphRAG's 100 facts in both settings, with reasoning 2 and 18 of them reaching the limit (Appendix~\ref{app:cwq}).}
At most 6.1 percent of a condition's calls reached the limit\supp{ (on the subsets of Appendix~\ref{app:reason} and under the status prompt, 75 of 16,420 calls limited to 256 tokens and 127 of 26,496 with reasoning)}, none on the validation questions.\supp{The 127 are counted less 20 calls repeated in a test run. GPT-5 and GPT-5-mini run at effort minimal with 128 tokens per step and 256 for the answer as the exploring LLM, and repair a contradicted context with 256, which no call reached; GPT-5 repairing with reasoning at effort low allows 2,048, which 56 of its 228 calls reached.}\supp{ GPT-5's repair run with reasoning stopped at its spending ceiling of 3 US dollars after 228 of its 528 contexts, all 228 with a reversed edge.} A call that reaches its limit returns no answer, is scored as wrong
and was not made again in the original runs; Appendix~\ref{app:caps} bounds these answers and reports the later reruns\supp{Why not made again: with no temperature or seed set, an answer need not repeat, so repeating only these calls would not equal a run with a larger limit.}. All completions are
cached; by the run logs the proprietary models cost about 170 US dollars at list price\supp{Of the about 170 US dollars, 36 went to the full PathQuestion and MetaQA sets, 21 to SubgraphRAG's top 10, 20 and 50 on WebQSP, 36 to GPT-5 with reasoning on CWQ (35.09 for the two contexts, 0.20 for a 60-prompt trial run and 0.40 for re-asking the capped answers) and 3.54 to GPT-5's re-judgment of the natural-error labels (Appendix~\ref{app:learned}).}.

\paragraph{Tests and intervals.} \looseness=-1 Paired tests are exact McNemar tests on exact match and Hit, Wilcoxon signed-rank or bootstrap tests on F1, and bootstrap or sign tests on scores, all Holm-adjusted over the LLMs or fact budgets compared. Brackets give 95 percent confidence intervals, by bootstrap in the text; intervals and tests by seed cluster on CWQ are marked, and table captions name their method.

\paragraph{Constants and a new graph.} \looseness=-1
No constant was set on test questions, though path counts became the default consistency check after an ablation on
the test probes (Appendix~\ref{app:learned}). The lexicon's lift of 1.3 lies on a plateau: fitted on 80 percent of
PathQuestion's training questions, lifts of 1.1 and 1.3 pass 78 and 79 percent of the other 20 percent (at most 300 per subset), with 95.2
percent of the derived answers exact, and lifts of 1.5 and 2.0 only 61 percent, with 91.2. WebQSP's ten neighbours and
vote threshold of 0.6, CWQ's settings (Appendix~\ref{app:cwq}) and the routing gates of both (Appendix~\ref{app:riskcov})
were chosen on validation questions; the gates were then confirmed on 500 WebQSP and 1,501 CWQ training questions held out
under a pre-specified plan, read by the seven open LLMs with the decoding above and 512 output tokens (Appendix~\ref{app:prereg}).
The learned consistency check's remaining settings (Appendix~\ref{app:learned}), the $P(\text{none})$ skip, the 99 percent single-value rule, the cap of 64 paths per fact and the size bins, are insensitive or set by hand (supplementary material),\supp{Skipping paths with $P(\text{none})$ of at least 0.9, 0.95 or 0.99 leaves every detection rate on CLUTRR and PathQuestion unchanged and moves the false alarms by at most 0.5 points. A relation is single-valued at 99 rather than 100 percent of its subjects so that a few noisy subjects do not disqualify it. The cap of 64 paths per fact bounds the cost: no fact of a CLUTRR or PathQuestion compact context has more than 2, and 1.2 percent of the facts judged in WebQSP's reach it. The size bins separate compact contexts, of up to 20 facts, from larger ones, which hold more paths.} and the false-alarm rate of 1 percent is an operating point to choose, not a fitted value (sensitivity tables and the
threshold sweep: the supplementary material). For a new graph: fit the mapping on training questions and set its lift or vote
threshold on held-out ones; count relations along random walks of the graph, train TransE on its facts and read off
its single-valued relations; calibrate on consistent contexts sampled from held-out parts of the graph, at the
false-alarm rate one accepts; and decide on validation questions whether the derived answer or the LLM answers where
the checks pass, and whether the LLM answers or the pipeline abstains where they fail. No step needs a contradicted
example from the new graph; for the default check, flagging a context by the smaller of its two shares and splitting the
threshold between them detect the same reversed facts built from training questions (Appendix~\ref{app:learned}).

\paragraph{Compact contexts.} \looseness=-1 With a learned mapping the selector keeps every path realising the predicted relation sequence, so it may reach an answer by two paths. A path fact whose deletion leaves the sequence's answers unchanged occurs in 6.4, 11.3 and 24.9 percent of the PathQuestion, WebQSP and MetaQA (lexicon) contexts where the sequence yields an answer (23.6 with MetaQA's classifier), and facts off every path (other facts of the paths' unnamed records, facts testing a constraint) in 36.8 percent of the WebQSP ones. CLUTRR's contexts are minimal: deleting any fact loses the derived label; 97 percent are simple paths, the rest pass a person twice where the rules compose no shorter walk.

\paragraph{Precise relation knowledge.} \looseness=-1 Table~\ref{app:setup:prec} measures, on questions not used for testing, how often the derived answer is exact where the derivability check passes; its third column names the question set. MetaQA's classifier is measured with its own and the lexicon's contexts, on both of which the check executes the classifier's sequence, and WebQSP's and CWQ's contexts are built as the test ones. PathQuestion's held-out questions also set the lexicon's lift, and the WebQSP and CWQ validation questions tuned $K$, the constraint share and $\tau$, so these values are not held out from tuning, which favours them. At a threshold of 90 percent, set post hoc, after the test results were known, CLUTRR's rules, MetaQA's classifier and PathQuestion's lexicon are precise and the chains of WebQSP and CWQ are not; any threshold above 85.0 and up to 95.2 percent gives the same split.

\begin{table*}[!t]
\centering
\scriptsize
\setlength{\tabcolsep}{4pt}

\begin{tabular}{l >{\raggedright\arraybackslash}p{36mm} >{\raggedright\arraybackslash}p{36mm} r r r}
\toprule
\textbf{benchmark} & \textbf{\begin{tabular}[b]{@{}c@{}}relation\\knowledge\end{tabular}} & \textbf{question set} & \textbf{questions} & \textbf{\begin{tabular}[b]{@{}c@{}}check passes\\(\%)\end{tabular}} & \textbf{\begin{tabular}[b]{@{}c@{}}derived answer\\right where it\\passes (\%)\end{tabular}} \\
\midrule
CLUTRR & rules & stories from the benchmark's generator, unused in testing & 257 & 100.0 & 100.0 \\
MetaQA & classifier, on the classifier's context & dev split, 300 sampled per hop, 886 kept & 886 & 100.0 & 99.9 \\
MetaQA & classifier, on the lexicon's context & dev split, 300 sampled per hop, 886 kept & 886 & 75.3 & 99.9 \\
PathQuestion & lexicon & held-out 20\% of the training questions, at most 300 per subset & 719 & 78.6 & 95.2 \\
WebQSP & chains predicted from similar training questions & validation split & 246 & 40.7 & 85.0 \\
CWQ & chains voted from the SPARQL queries of similar training questions & validation split & 3,519 & 9.9 & 61.6 \\
\bottomrule
\end{tabular}

\caption{Relation knowledge on questions not used for testing: the question set and its size, the share that pass the derivability check and the share of these whose derived answer is exact (\%).}
\label{app:setup:prec}
\end{table*}

\section{CLUTRR: added and shuffled facts}
\label{app:clutrr}

\looseness=-1 Plots of accuracy against the number of added facts and bootstrap intervals for Section~\ref{sec:inputs}'s conditions: the supplementary material. Under the forced-choice prompt the compact context reads more accurately than all facts of the story for six LLMs (exact McNemar, Holm-adjusted $p \leq 2.5\times10^{-6}$; EXAONE-3.5-7.8B: 0.6 points less, $p = 0.20$) and than as many facts most similar to the question for all seven ($p \leq 1.2\times10^{-9}$). Over the story the six LLMs' lead is 2.1 to 5.2 points; over the similar facts, which are sufficient for only 47.9 percent of questions (2,091 of 4,361; the selector table of the supplementary material), it is 3.6 to 21.0 points, and on the questions where they are sufficient the two contexts read within 0.6 points of each other for every LLM. Under the forced-choice prompt one reversed-edge contradiction costs 31.8 to 42.4 percent of the compact context's accuracy.

\paragraph{Order held fixed.} \looseness=-1 Under the abstention prompt two distractors attached to the path cost 19 to 56 percent of the compact context's accuracy; under the forced-choice prompt shuffling its facts costs 5 to 25 points ($p<10^{-15}$). Our context lists its facts in path order, the story in story order; with both read shuffled under the abstention prompt (the supplementary material), the compact context still reads significantly better (Holm-adjusted $p < 0.02$) for five LLMs, by 1.6 to 5.1 points, while EXAONE-3.5-7.8B gains 1.3 and Llama-3.2-3B loses 1.0 (Holm-adjusted $p = 0.10$ each).

\section{Ambiguous compositions on CLUTRR}
\label{app:depth}

\paragraph{Questions whose label real kinship leaves open.} \looseness=-1 The 2,307 unambiguous questions avoid the four step and in-law compositions; 330 of them, the strict subset, also avoid composing siblings through a shared parent, which a half-sibling would break, and are all two-fact chains. The derived answer is right for 2,306 of the 2,307 unambiguous questions and all 330 strict ones. Under the forced-choice prompt, six LLMs read the compact context 3.2 to 6.1 points more accurately than all facts of the story on the unambiguous questions and EXAONE-3.5-7.8B 0.6 points less; on the strict subset five read it 2.1 to 7.0 points more accurately, significantly for two (Holm-adjusted exact McNemar $p \leq 0.004$), Gemma-3-27B reads both perfectly and EXAONE-3.5-7.8B loses 7.6 points (Holm-adjusted $p = 0.005$). The three largest LLMs abstain on the strict subset's depleted contexts themselves, so gating on the rules' derivability check gains them at most 2 points of balanced score, while Llama-3.1-8B and EXAONE-3.5-7.8B still answer 68 and 64 percent of them and gain 32 to 34. Per LLM, with and without the compositions of Appendix~\ref{app:setup}, by chain length and fact-set size, and all three subsets per comparison: the supplementary material.

\section{PathQuestion and MetaQA}
\label{app:realkb}

\looseness=-1 On PathQuestion with the forced-choice prompt, two controls built without relation knowledge, a random path of the gold length from the topic entity and a path along relations the LLM itself names before the facts are selected, trail the compact context by 23.1 to 39.1 and 12.4 to 32.5 exact-match points for every LLM (Holm-adjusted exact McNemar $p \leq 7\times10^{-20}$). Per LLM with Hit and F1, MetaQA, the annotated path by number of answers, gating, truncation and versions of MetaQA's derivability check: the supplementary material.

\section{Published baselines}
\label{app:baselines}

\looseness=-1 Every published method is read with the same forced-choice prompt, LLMs and exact-match scoring as our contexts; only the facts differ: KAPING \citep{baek2023} gives the 10 incident triples most similar to the question, SubgraphRAG \citep{li2025} its released top-100 triples on WebQSP and the 100 highest-scoring of its retriever retrained with the official code on PathQuestion and MetaQA (50 on average on MetaQA, whose question graphs may hold fewer), Think-on-Graph \citep{sun2024} the triples of the reasoning chains of a port of its official implementation with its defaults, and RoG \citep{luo2024} its released planned paths instantiated in the question's graph (the settings in full: supplementary material). RoG's paths on CWQ are read likewise (Appendix~\ref{app:cwq}). The published end-to-end results of these systems use other LLMs and prompts, such as RoG's fine-tuned model or SubgraphRAG's own prompt, and are not comparable: we compare the contexts these methods select, read by the same LLMs, not their full pipelines. Over the seven LLMs the compact context leads the context selected by KAPING in exact match by 7.6 to 34.7 points on CLUTRR and 39.3 to 67.8, 20.7 to 38.0 and 22.4 to 28.6 on PathQuestion, MetaQA and WebQSP, SubgraphRAG by 15.1 to 26.6, 6.9 to 13.2 and 4.6 to 13.0, Think-on-Graph by 23.4 to 34.1, 20.4 to 32.6 and 22.1 to 29.8, and RoG on WebQSP by 14.9 to 18.3 (exact McNemar $p \leq 7\times10^{-6}$; against SubgraphRAG on WebQSP $p \leq 3\times10^{-4}$). Per question, ours takes one call of 152 to 452 prompt tokens (171 to 452 without CLUTRR), KAPING 140 to 343, SubgraphRAG 951 to 2,313 (50.3 to 100.0 facts on average), RoG 478 to 565, and Think-on-Graph 2,605 to 6,558 over 5.8 to 11.4 calls. The supplementary material gives the summary and per-LLM tables and how Think-on-Graph's search ended.

\paragraph{Fact order.} \looseness=-1 Our contexts list their facts in path order, the published methods in their own. Shuffled, with the same prompt and output limits, ours lose 0.2 to 13.7 points of exact match on PathQuestion, MetaQA and WebQSP and stay ahead of every published method's context for every LLM, significantly except over SubgraphRAG for Qwen3-32B on WebQSP ($p = 0.31$). On CLUTRR they stay ahead of KAPING's context for five LLMs (Holm-adjusted exact McNemar $p \leq 0.004$), not for Qwen2.5-7B ($+0.7$, $p = 0.35$), and fall 2.1 points behind for Llama-3.2-3B ($p = 0.002$).

\paragraph{SubgraphRAG at every budget and in path order.} \looseness=-1 The supplementary material cuts SubgraphRAG's ranked facts, in retrieval order, to our context's size per question (median 2, 3 and 4 facts on PathQuestion, MetaQA and WebQSP) and to 10, 20 and 50 besides its full 100 (a question with fewer facts keeps all); path order changes the exact match of SubgraphRAG's 100 WebQSP facts by $-2.5$ to $+1.3$ points. Our contexts hold every gold answer for 97.3, 77.3 and 72.9 percent of questions in the same order, about as often as SubgraphRAG's 100 facts where its retriever was retrained (100.0 and 73.2) and less often on WebQSP (84.6). At every budget all seven LLMs read our context more accurately (Holm-adjusted $p \leq 2.5\times10^{-4}$), and at the budget each LLM reads best it stays 10.2 to 19.9, 5.9 to 12.7 and 4.6 to 13.0 points ahead in the same order; on PathQuestion, SubgraphRAG's 20 facts hold every gold answer for 99.7 percent of questions yet trail by 10.2 to 22.5 points, so recall does not explain the gain.

\paragraph{SubgraphRAG's own prompt.} \looseness=-1 SubgraphRAG reports a macro-F1 of 70.6 and a Hit of 86.6 on WebQSP with Llama-3.1-8B and its own prompt \citep{li2025}; with our forced-choice prompt and 4-bit weights Llama-3.1-8B reads the same facts at 62.7 and 79.2. The supplementary material reads both WebQSP contexts with that prompt, scoring replies as in our runs and with SubgraphRAG's evaluators. Under its evaluators Llama-3.1-8B reads SubgraphRAG's facts at 67.1 and 84.3, near the published values (61.1 and 75.2 under ours). Our context stays ahead in exact match for five LLMs, by 6.0 to 12.2 points (Holm-adjusted $p \leq 7\times10^{-6}$), but not for Qwen3-32B and EXAONE-3.5-32B (0.1 and 1.4, $p \geq 0.49$), and SubgraphRAG's facts give every LLM more hits, by 2.8 to 11.6 points under its evaluators. Against the forced-choice prompt every LLM loses 3.5 to 15.7 points of exact match on our context, against $-10.5$ to $+2.8$ on SubgraphRAG's facts; up to 12.2 percent of replies give no answer.

\begin{figure*}[!t]
\centering
\includegraphics[width=\textwidth]{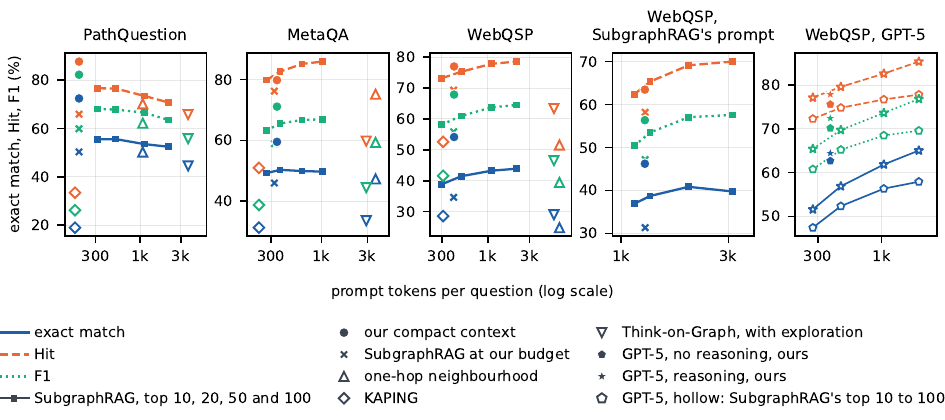}
\caption{Mean exact match (blue, solid), Hit (orange, dashed) and F1 (green, dotted) against mean prompt tokens per question (log scale; each LLM's tokenizer, GPT-5's as billed); lines join SubgraphRAG's 10, 20, 50 and 100 facts, other markers as in the legend. Panels 1 to 3: forced-choice prompt, the seven open LLMs (six for WebQSP's one-hop neighbourhood, Appendix~\ref{app:setup}). Panel 4: WebQSP, SubgraphRAG's own prompt, the same LLMs; its tokens include SubgraphRAG's second request for the answer format, sent for 4.5 to 8.0 percent of questions on average. Panel 5: WebQSP, GPT-5, forced-choice prompt; pentagons without reasoning, stars with it; our context filled, SubgraphRAG's facts hollow.}
\label{app:baselines:cost}
\end{figure*}

\paragraph{Accuracy against prompt tokens.} \looseness=-1 With the forced-choice prompt and SubgraphRAG's facts cut to our context's number of facts or to 10, 20, 50 or 100 facts (Figure~\ref{app:baselines:cost}), every open LLM scores lower than on our context in exact match (Holm-adjusted $p<0.05$ throughout) and in F1, where the difference is significant by Holm-adjusted bootstrap tests for only 4 of the seven LLMs at 50 and 100 facts on MetaQA and 6 at 50 and 4 at 100 facts on WebQSP. Hit averaged over the LLMs reaches ours at 0.78 times our prompt tokens on MetaQA (10 facts; at 100 facts significantly higher for six LLMs), 2.68 times on WebQSP (50 facts) and at no budget on PathQuestion.

\section{WebQSP}
\label{app:webqsp}

\paragraph{Relation knowledge from predicted chains.} \looseness=-1 WebQSP annotates each training question with its inferential chain, the relations from the topic entity to the answer, and with constraints: equalities, dates relative to the present and orderings; a date named in a neighbour's own question cannot be carried over. A test question, its topic mention removed, is embedded with MPNet (all-mpnet-base-v2), and its $K$ most similar training questions each vote for their chain with their cosine similarity. Among the chains that can be followed from the topic entity in the question's graph we take the one with most votes; where none can (97 test questions), the context is empty and the check fails. The context holds every path that realises the chain, the other facts of the unnamed nodes on those paths (Freebase's records grouping the facts of an event, such as a marriage or a term of office), and the facts that test each constraint shared by at least half of the voters; the derived answer is the set of path ends satisfying them. The derivability check passes when the chain holds at least a share $\tau$ of the votes, every shared constraint can be tested on the context, and some path end satisfies all of them. On the 246 validation questions, $K=10$ was chosen among 5, 10 and 20 by the derived answer's exact match, and $\tau=0.6$ is the smallest threshold whose derived answer is exact for at least 80 percent of the questions that pass (85.0 percent, with 40.7 percent passing, on the questions that chose it). The graphs keep no dates, so the check fails where the vote carries a date or ordering constraint; it passes 54 of the 143 test questions annotated with one, whose vote lacks it.

\paragraph{Weak supervision and the lexicon.} \looseness=-1 The weakly supervised variant predicts instead from the directed relation sequences of the shortest question-to-answer paths in the training graphs, computed with RoG's own code, without constraints ($\tau=0.5$ by the same rule). The lexicon selector applies the PathQuestion lexicon, fitted on the same paths, to paths of one or two steps from the topic entity. In the supplementary tables, kNN chains, the chains voted by the $k$ nearest neighbours, are those predicted from the ten most similar training questions, and every check reads the compact context itself.

\paragraph{Reference contexts.} \looseness=-1 Two contexts complete Table~\ref{tab:inputs}: as many of the question graph's facts most similar to the question as the kNN chains' context holds, and the question's own annotated chain and constraints, empty for the 125 questions whose graph does not realise that chain.

\looseness=-1 The kNN chains' contexts hold 14.7 facts on average and add 6.0 to 7.6 exact-match points over the weakly supervised chains', which beat RoG's paths in exact match for every LLM, by 7.7 to 11.6 points, and SubgraphRAG's 100 facts for six (Holm-adjusted exact McNemar $p<0.05$ for four), with fewer hits than both for every LLM. Per LLM and context, with routing: the supplementary material.

\paragraph{How far the WebQSP results reach.} \looseness=-1 \label{app:webqsp:scope}Of the 1,628 test questions, 1,542 (94.7 percent) have an annotated chain among the 392 training chains from which chains are predicted, 63 have only unseen chains and 23 none; the derivability check passes 47.7 and 6.3 percent of the first two groups, with the derived answer right for 79.5 percent and 0 of 4 of those passing. Of the questions whose shortest annotated chain has one relation (1,032) or two (573), 50.0 and 39.1 percent pass, with the derived answer right for 87.0 and 60.7 percent of those passing. WebQSP thus has few test questions with unseen chains and no annotated chain longer than two relations. Over all test questions, passing or not, the derived answer reaches 64.4 F1 and routing 68.4 to 72.1 (seven open LLMs); semantic-parsing methods that use the full Freebase graph, scored on all 1,639 test questions, report 72.8 (CBR-KBQA), 75.6 (RnG-KBQA as first reported; 76.2 rerun by DecAF with T5-large), 77.1 and 78.8 (DecAF with FiD-large and FiD-3B), all from Table~1 of \citet{yu2023decaf}. Of the policies computed from existing outputs, reading SubgraphRAG's facts where the check fails changes routing's exact match by $-4.3$ to $+0.7$ points and choosing, where it passes, whichever answer is right (an oracle) adds at most 2.6; the LLM under the abstention prompt where the check fails scores $+12.7$ to $+35.7$ per 100 questions against $+26.2$ for the derived answer or an abstention, and behind WebQSP's routing gate routing reaches 68.1 to 70.3 F1 (in full: the supplementary material).

\section{ComplexWebQuestions}
\label{app:cwq}

\paragraph{Data and relation knowledge.} \looseness=-1 CWQ extends WebQSP queries by composition, conjunction, a comparison or a superlative \citep{talmor2018}. We read RoG's question graphs for CWQ's 27,639 training, 3,519 validation and 3,531 test questions; for 683 test questions (19.3 percent) the graph reaches no gold answer. Each SPARQL query is parsed into a start entity, a relation chain to the answer and constraints on the chain's nodes (an entity reached by one or two relations, a date, a comparison or ordering on a literal), for 99.0 and 98.2 percent of training and test queries. A constraint value among the question's entities is kept as such; another is named only where its training occurrences agree. As on WebQSP, the 20 training questions most similar under MPNet vote for a chain; constraints shared by at least 30 percent of the voters are applied, and the check passes when the chain holds at least 80 percent of the votes, every applied constraint can be tested (an unnamed value cannot; 270 test questions apply one) and some path end meets them; where no voted chain can be followed in the question's graph, our context is empty (637 test questions, 18.0 percent). $K$, the share and the vote threshold were chosen on all validation questions, where 9.9 percent pass with 61.6 percent exact (the grid: supplementary material). The graphs keep no literals, so comparisons, superlatives and dates cannot be tested; a comparative or superlative question passes only when none is applied. The weakly supervised variant votes for RoG's shortest question-to-answer paths ($K=10$, threshold 0.3, chosen alike). SubgraphRAG's 100 facts and RoG's paths come from the scored triples and predictions released for CWQ, and the reference executes the test question's own parsed query on its graph (empty for 787). Table~\ref{tab:inputs}'s one-hop neighbourhood, similarity top-$k$ and Think-on-Graph were not run on CWQ, for compute; SubgraphRAG's facts cut to the size of ours (at least one fact, so 18.8 against 18.6 on average) serve as the size-matched relevance control instead, the stronger of the two on WebQSP (34.6 against 21.6 percent exact).

\paragraph{Results.} \looseness=-1 The check passes 417 test questions (11.8 percent); the derived answer is exact for 52.0 percent of them and, used also where the check fails, 22.3 percent of all (Hit 51.2, F1 33.1). The supplementary material compares the contexts per LLM. Over the seven open LLMs, with mean exact match, Hit and F1 per context and, where stated, ours minus the context in exact match (for SubgraphRAG's 100 facts also in Hit and F1) as a range with the LLMs with ours ahead / behind at Holm-adjusted $p<0.05$: compact (ours) 27.3 / 52.0 / 40.7; SubgraphRAG, 100 facts 27.0 / 54.3 / 41.5, ours minus it $-$3.3 to +3.3 (2/1) / $-$4.9 to +2.5 (1/4) / $-$2.4 to +2.4 (2/5); RoG 22.4 / 48.6 / 36.7, ours minus it +2.9 to +6.4 (7/0). Table~\ref{app:passfail:split} compares our context with SubgraphRAG's 100 facts where the check passes. By composition type (supplementary material), the check passes for 5.8, 19.9, 3.3 and 3.0 percent of the 1,546 composition, 1,575 conjunction, 213 comparative and 197 superlative questions, the derived answer is exact for 77.8, 46.8, 0.0 and 0.0 percent of those (90, 314, 7 and 6). Our context takes 505 prompt tokens on average over the LLMs, SubgraphRAG's 100 facts 2,067. Where the check passes, the derived answer is right for 52.0 percent and the LLMs for 24.9 to 60.4; it is significantly above the four smallest LLMs and below the three largest (Holm-adjusted $p<0.05$): CWQ's relation knowledge is not precise by the 90 percent line, as its validation precision showed. Behind the one-answer gate chosen on validation, the derived answer is right far more often (below).

\paragraph{GPT-5 without and with reasoning.} \looseness=-1 GPT-5 read every CWQ test question with our context and with SubgraphRAG's 100 facts, without reasoning (effort minimal, output limit 256 tokens; 0 and 2 answers cut at the limit; USD 11.74 at list price) and with reasoning (effort low, limit 2,048; 2 and 18 cut; USD 35.09; tables: the supplementary material). On all questions SubgraphRAG's facts are read better, more so with reasoning: exact match 44.7 against 41.3 without reasoning ($-$3.5 points, $p = 1.0\times10^{-5}$) and 51.2 against 43.6 with it ($-$7.6, $p = 6.1\times10^{-25}$), and so in Hit and F1; reasoning raises GPT-5's exact match by +6.5 points on SubgraphRAG's facts and +2.4 on ours, while the seven open LLMs read the two within 3.3 points of each other (Results, above). GPT-5 has no CWQ validation run. Re-asked once at 8,192 tokens, all 20 answers capped with reasoning finished (5 exact; USD 0.40), moving no difference by more than 0.5 points and no verdict; the worst case for the stored capped answers changes no verdict either.
\paragraph{Qwen3-32B with reasoning.} \looseness=-1 Qwen3-32B with reasoning (the model of the other Qwen3-32B runs, output limit 4,096 tokens, answers scored as GPT-5's) read the same prompts (tables: the supplementary material). SubgraphRAG's facts are again read better, by more: exact match 44.4 against 38.7 ($-$5.7 points, $p = 6.7\times10^{-13}$), Hit 56.5 against 48.4, F1 51.1 against 43.9 ($p = 3.6\times10^{-21}$). Of its 220 answers at the output limit, none exact, the worst case keeps 19 of 21 verdicts (in full: the supplementary material).
\paragraph{The abstention prompt on CWQ.} \looseness=-1 Under the abstention prompt (the forced-choice prompt with the line permitting the reply INSUFFICIENT\_CONTEXT), the seven open LLMs answer 41.5 to 62.6 percent of the questions on our context and 66.2 to 84.8 on SubgraphRAG's 100 facts, and offering abstention costs 1.8 to 8.0 and 0.5 to 5.7 points of exact match; the abstentions are poorly placed (19.7 to 40.3 percent of the contexts holding every gold answer refused, 13.7 to 35.4 percent of those lacking one answered).

\paragraph{The one-answer gate.} \looseness=-1 On CWQ the size of the derived answer set is a better confidence signal than the check (Appendix~\ref{app:riskcov}), so CWQ's routing gate is chosen on its 3,519 validation questions among rules on the check and the set size, by the open LLMs' mean routed exact match: `the derived answer set holds one answer' passes 24.6 percent of the validation and 24.5 of the test questions (864), with the derived answer exact for 69.2 and 68.6 percent, against 417 questions and 52.0 percent for the check. Where it passes, the derived answer is above the four smallest LLMs (Holm-adjusted $p \leq 8.9\times10^{-31}$) and level with the three largest and the strong LLMs (66.3 to 70.3 percent, $p \geq 0.09$). Routing behind the one-answer gate raises every open LLM's exact match over the LLM alone, by +0.2 to +8.9 points (four significantly), and over routing behind the check by +0.7 to +5.7 (all seven significantly), at a cost of 0.3 to 1.1 points of Hit. Passing questions with one derived answer are wrong for 3.6 percent, those with two or more for 86.5 (223 questions). The precision gate, `the check passes and one answer is derived', named after the test precision below had been seen (its held-out precision, Appendix~\ref{app:prereg}, is the confirmatory number), passes 194 test questions (5.5 percent; validation 5.4, 96.8 percent exact), the derived answer exact for 96.4 percent (Hit 97.4, F1 96.8), above the open LLMs there at 42.3 to 90.7 percent (Holm-adjusted $p \leq 0.002$) and Qwen3-32B with reasoning at 91.2 ($p = 0.006$; by seed cluster 0.016), and level with or above GPT-5 at 93.3 without and with reasoning (McNemar $p = 0.031$ each; by seed cluster 0.06 and 0.12).
\paragraph{Who answers, chosen on validation.} \looseness=-1 As on WebQSP (Appendix~\ref{app:riskcov}), validation questions choose per open LLM among six policies, the four fixed ones of Appendix~\ref{app:routing} and the LLM under the abstention prompt, alone or behind the derived answer where the routing gate passes (the open LLMs' validation reads under that prompt), by accuracy and by the equal-cost score (Section~\ref{sec:who}: $+1$ per right answer, $-$1 per wrong one, 0 per abstention). By accuracy they pick routing for five open LLMs and the LLM alone for Qwen3-32B and EXAONE-3.5-32B; by the score, the derived answer or an abstention for the four smallest and the LLM's own abstention for the three largest; validation overstates the chosen scores by 0.3 to 4.8 points. GPT-5 and Qwen3-32B with reasoning have no CWQ validation reads, so no policy is chosen for them (in full: the supplementary material).

\section{Where the routing gate passes}
\label{app:passfail}

\tabdecomp
\paragraph{What each part contributes on WebQSP.} \looseness=-1 Table~\ref{tab:decomp} separates the parts on WebQSP with the chain predictor fixed. Our context instead of SubgraphRAG's facts gains the open LLMs 10.2 points of mean exact match with annotated chains but 3.6 with weakly supervised ones, whose own contexts read 6.7 points worse than the annotated chains' contexts. Routing to the derived answer where the routing gate passes adds 5.0 points, significantly for the four smallest LLMs. Two control variants move the open LLMs' mean exact match under routing by less than one point: picking as many questions by the vote share instead of the routing gate, and also requiring the consistency check to find no conflict. On WebQSP the gain thus comes from selecting the context and from returning the derived answer; whether the gate or the vote share picks the questions that get the derived answer matters little.

\looseness=-1 Table~\ref{app:passfail:split} compares our compact context with SubgraphRAG's 100 facts where the routing gate on our context passes: the derivability check on PathQuestion and MetaQA, on WebQSP the check or at most 20 derived answers and on CWQ one derived answer, rules chosen on validation questions (Section~\ref{sec:checks}; a setting is one LLM reading one benchmark under one prompt). The rows above the pooled row are the 53 settings of Section~\ref{sec:inputs}; below it, CWQ is split by the derivability check instead, and the last row is a supervision control; the supplementary material gives the same table with the failing side. Where the routing gate passes, our context is significantly ahead ($p<0.05$) for 52 of the 53 settings in exact match (behind for 0), 44 in F1 (behind for 2) and 36 in Hit (behind for 6); where it fails, SubgraphRAG's facts give significantly more hits for 23 of the 34 settings with failing questions, CWQ's one-answer gate being split on its passing side only. In the WebQSP and MetaQA lexicon rows, failing questions contribute $-18.6$ to $-0.2$ points of Hit to the difference over all questions and passing questions $-6.9$ to $+8.8$ (on CWQ under the check, $-4.8$ to $+1.4$ and $-0.6$ to $+1.1$), so SubgraphRAG's Hit lead comes mostly from failing questions. The split follows our pipeline's verdict, so the subsets differ in difficulty: SubgraphRAG's exact match is higher on passing than on failing questions for all 32 WebQSP and MetaQA lexicon settings, the 7 with weakly supervised chains included, all 7 CWQ settings under the one-answer gate and all 7 under the check, and for none of 9 on PathQuestion. The table shows where each context reads better, not a randomised effect of the routing gate.

\begin{table*}[!t]
\centering
\scriptsize
\setlength{\tabcolsep}{5pt}
\begin{tabular}{@{}l r r r@{}}
\toprule
 & \multicolumn{3}{c}{\textbf{routing gate passes: ours minus SubgraphRAG's 100 facts}} \\
\cmidrule(lr){2-4}
\textbf{setting} & \textbf{exact match} & \textbf{Hit} & \textbf{F1} \\
\midrule
WebQSP & +5.5 to +15.1 (7/0) & $-$1.3 to +3.5 (1/0) & +1.0 to +9.4 (6/0) \\
WebQSP, SubgraphRAG's prompt & +3.2 to +16.9 (7/0) & $-$8.2 to $-$2.0 (0/5) & $-$2.7 to +4.3 (2/1) \\
WebQSP, GPT-5 without reasoning & +6.8 (1/0) & +0.9 (0/0) & +2.9 (1/0) \\
WebQSP, GPT-5 with reasoning & +1.2 (0/0) & $-$4.4 (0/1) & $-$2.2 (0/1) \\
CWQ, one-answer gate & +8.1 to +16.2 (7/0) & +4.4 to +13.0 (7/0) & +8.3 to +12.0 (7/0) \\
CWQ, one-answer gate, GPT-5 without reasoning & +6.2 (1/0) & +4.7 (1/0) & +5.2 (1/0) \\
CWQ, one-answer gate, GPT-5 with reasoning & +2.9 (1/0) & +1.2 (0/0) & +1.7 (0/0) \\
CWQ, one-answer gate, Qwen3-32B reasoning & +5.2 (1/0) & +0.7 (0/0) & +2.4 (0/0) \\
MetaQA, lexicon context & +12.8 to +27.5 (7/0) & +5.5 to +9.1 (7/0) & +14.8 to +20.7 (7/0) \\
MetaQA, classifier context & +16.7 to +32.2 (7/0) & +8.4 to +12.3 (7/0) & +17.9 to +24.5 (7/0) \\
PathQuestion & +18.9 to +30.7 (7/0) & +11.8 to +29.6 (7/0) & +15.1 to +27.1 (7/0) \\
PathQuestion, GPT-5 reasoning & +7.1 (1/0) & +5.1 (1/0) & +5.9 (1/0) \\
MetaQA, lexicon, GPT-5 reasoning & +26.4 (1/0) & +6.6 (1/0) & +18.9 (1/0) \\
MetaQA, classifier, GPT-5 reasoning & +30.0 (1/0) & +7.7 (1/0) & +21.3 (1/0) \\
PathQuestion, Qwen3-32B reasoning & +8.6 (1/0) & +5.0 (1/0) & +6.5 (1/0) \\
MetaQA, lexicon, Qwen3-32B reasoning & +29.0 (1/0) & +11.4 (1/0) & +22.9 (1/0) \\
MetaQA, classifier, Qwen3-32B reasoning & +34.4 (1/0) & +11.4 (1/0) & +25.0 (1/0) \\
\midrule
pooled: 53 settings above & +1.2 to +34.4 (52/0) & $-$8.2 to +29.6 (36/6) & $-$2.7 to +27.1 (44/2) \\
\midrule
CWQ, derivability check & +5.3 to +15.3 (7/0) & $-$5.0 to +9.4 (1/0) & +8.2 to +11.0 (7/0) \\
WebQSP, weakly supervised chains & +1.2 to +15.3 (6/0) & $-$2.9 to +1.4 (0/0) & +0.4 to +6.3 (6/0) \\
\bottomrule
\end{tabular}
\caption{Our context minus SubgraphRAG's 100 facts (points) where the routing gate passes: the derivability check on PathQuestion and MetaQA, the check or at most 20 derived answers on WebQSP and one derived answer on CWQ, rules chosen on validation questions; the pooled row covers the 53 settings of Section~\ref{sec:inputs}. Below it, CWQ split by the derivability check instead, and a supervision control split by its own check. Forced-choice prompt unless SubgraphRAG's is named; under SubgraphRAG's prompt its evaluators score exact match, Hit and macro-F1. Cells: range over the row's LLMs, the seven open LLMs or the named LLM, and in parentheses those with ours / SubgraphRAG's ahead at $p<0.05$ (Appendix~\ref{app:setup}'s tests, Holm-adjusted over a row's seven open LLMs; F1: Wilcoxon signed-rank). Passing: WebQSP 1,385 of 1,628 (85.1\%; weakly supervised chains 417, 25.6), CWQ 864 of 3,531 by the one-answer gate (24.5) and 417 by the check (11.8), MetaQA 694 and 900 of 900 (lexicon, classifier; 77.1, 100.0), PathQuestion 602 of 782 (77.0). The failing side: the supplementary material.}
\label{app:passfail:split}
\end{table*}

\paragraph{A combination guided by the routing gate.} \looseness=-1 After seeing both results (post hoc), we defined a combination: the LLM reads our context where the routing gate passes, else SubgraphRAG's facts. It has no tuned parameter but was not confirmed on validation questions, which no LLM read with SubgraphRAG's facts. On WebQSP it beats SubgraphRAG's facts for every open LLM in exact match, by 4.7 to 12.8 points, at 29 to 30 percent of SubgraphRAG's mean prompt tokens. With the derived answer where the routing gate passes, the WebQSP gains are 3.9 to 25.3 points of exact match (all seven significant), $-0.6$ to $+5.5$ of Hit (2) and $+0.3$ to $+12.4$ of F1 (5) at 15 percent of the tokens.

\paragraph{Relation supervision fixed.} \looseness=-1 With one predicted chain per question for every policy (WebQSP, forced-choice prompt), the seven open LLMs' mean exact match gains 10.2 points over SubgraphRAG's facts with our context (selection), 10.4 with the combination; routing adds 5.0 to our context (Holm-adjusted $p<0.05$ for 4 of 7 LLMs). Replacing the routing gate by a vote threshold returning as many derived answers, or adding either version of the consistency check, changes each LLM's exact match under routing and both combinations by $-0.4$ to $+0.6$ (smallest Holm-adjusted $p = 0.11$). Annotated instead of weakly supervised chains raise mean exact match by 6.7 points with our context, 11.3 with routing.

\paragraph{Budget-matched controls.}\label{app:passregion} \looseness=-1 Where the routing gate passes, SubgraphRAG's facts cut to our context's size (the same ranked list as its 100 facts) read worse than our context for all seven open LLMs in every setting with that run (on WebQSP by +19.8 to +25.0 points of exact match under the forced-choice prompt), in exact match, Hit and F1, significantly (Holm-adjusted $p<0.05$) except in Hit on CWQ for 3 of the seven (the supplementary material; the test and adjustment of Table~\ref{app:passfail:split}), and by more than its 100 facts do, and the cut list is less accurate there (mean exact match over the LLMs, the cut list against 100 facts: WebQSP 37.3 against 47.0, CWQ 27.9 against 33.2; ours 59.3 and 45.6). The same holds in exact match against its 10 and 20 facts (with the failing questions: the supplementary material).
\paragraph{SubgraphRAG's own high-confidence region.} \looseness=-1 A high-confidence region chosen by SubgraphRAG itself, the $n$ questions with the highest top-1 triple score where $n$ is our pass count, overlaps our passing region for 1,201 of 1,385 WebQSP, 78 of 417 CWQ, 580 of 694 MetaQA (lexicon) and 474 of 602 PathQuestion questions (the supplementary material; expected under independence 1,178, 49, 535 and 463); on WebQSP, where the routing gate passes 1,385 questions, the two regions nearly coincide (Jaccard 0.77), so there this control no longer separates them. Within SubgraphRAG's own region our context still leads in exact match for all seven open LLMs on WebQSP, MetaQA and PathQuestion and for GPT-5 without reasoning on WebQSP ($-$0.3 to +5.0 over GPT-5 without and with reasoning), but for two of seven on CWQ ($-$2.4 to +9.8), where our lead sits in the passing questions SubgraphRAG scores low. Selecting on the retriever's confidence does not reproduce the passing-region lead; on CWQ it reproduces part of it.
\paragraph{Stratified by SubgraphRAG's confidence.} \looseness=-1 Within the passing region, split by quartiles of SubgraphRAG's top-1 score over the whole test set, our context is ahead in every quartile on MetaQA and PathQuestion (6 to 7 of the seven open LLMs significant per quartile), on WebQSP in the three upper quartiles (6 to 7; GPT-5 in the two upper ones) and in the lowest by $-$0.6 to +12.6, and ahead or level for every LLM in every CWQ quartile (4, 4, 4, 1 of 7 significant; the supplementary material).
\paragraph{A cross-fitted policy.} \looseness=-1 The combination policies of Appendix~\ref{app:passfail} were defined after the test results were seen, and so was the family of five policies below; no validation run can confirm them, as no LLM read SubgraphRAG's facts on validation questions. Cross-fitting on the test questions, under a protocol written down before its numbers were computed, therefore measures how stable a choice within that family is across halves, not an independent confirmation. The derived-answer combination is chosen in every fold on MetaQA and PathQuestion, with out-of-fold gains of +14.9 to +30.0 (lexicon), +10.9 to +41.4 (classifier) and +4.7 to +28.9 points over the seven open LLMs and +2.9 to +13.0 for the six reasoning settings. On the human-written benchmarks the gain over the better single context is small or unstable: WebQSP $-$0.5 to +12.3, CWQ $-$0.3 to +3.2.
\paragraph{Effect sizes and units.} \looseness=-1 Each setting is one LLM reading one benchmark and context under one prompt: the 53 settings are the rows of Table~\ref{app:passfail:split} above its pooled row, CWQ by the one-answer gate with the three full CWQ runs of Appendix~\ref{app:cwq} (GPT-5 without and with reasoning, Qwen3-32B with reasoning). With the passing region the routing gate (on WebQSP and CWQ the rule chosen on validation questions, on PathQuestion and MetaQA the check), our context leads SubgraphRAG's 100 facts in exact match in all 53, significantly in 52 (not GPT-5 with reasoning on WebQSP, +1.2, $p = 0.33$), by a median of +15.1 points (quartiles +9.7 and +20.0, range +1.2 to +34.4; on CWQ +2.9 to +16.2); in F1 in 50 (44 significantly; median +11.7), significantly behind in 2, GPT-5 with reasoning on WebQSP ($-$2.2, $p = 0.029$) and EXAONE-3.5-32B under SubgraphRAG's prompt ($-$2.7, Holm-adjusted $p = 0.043$); in Hit in 41 (36 significantly, 6 significantly behind). The settings are not independent, so LLMs serve as units for a sensitivity check (a unit improves only if every one of its settings does; two-sided sign test): 9 of 9 LLMs improve in exact match ($p = 0.0039$), 8 of 8 with Qwen3-32B with and without reasoning merged ($p = 0.0078$) and 6 of 6 model series ($p = 0.031$); in F1 6 of 9 ($p = 0.51$). With the derivability check as CWQ's region instead, 51 of 53 settings lead in exact match and 48 in F1: Qwen3-32B with reasoning trails by 1.7 points (significantly in Hit and F1) and GPT-5 with reasoning by 0.7 ($p = 0.83$), so 7 of 9 units improve ($p = 0.18$).

\paragraph{Weak supervision against SubgraphRAG.} \looseness=-1 Chains voted from RoG's shortest question-to-answer paths of similar training questions (weak supervision; Appendix~\ref{app:webqsp}) replace annotated parses. Against SubgraphRAG's 100 facts, the seven open LLMs read them on WebQSP in exact match at 47.4 against 43.9 (4 ahead, 0 behind), but in Hit at 72.2 against 78.6 (0 ahead, 7 behind) and in F1 at 61.9 against 64.5 (0 ahead, 3 behind); on CWQ at 22.5 against 27.0 (0 ahead, 5 behind), 45.1 against 54.3 (0 ahead, 7 behind) and 34.1 against 41.5 (0 ahead, 7 behind) (means of the seven; LLMs significantly ahead or behind, Holm-adjusted). The selection gain over SubgraphRAG thus needs annotated parses; with weak supervision it holds only for exact match on WebQSP.

\section{Gating on CLUTRR}
\label{app:gating}

\looseness=-1 For the two probes that make a compact context insufficient, the supplementary material gives each LLM's answer rate and balanced score alone and behind each gate, including the learned derivability check of Appendix~\ref{app:learned}, and Table~\ref{tab:faith} the balanced scores with the rules and the three-way scores with the rules and the learned checks. With a necessary fact deleted, Llama-3.2-3B, Llama-3.1-8B and EXAONE-3.5-7.8B still answer 57.6 to 61.3 percent of contexts and Qwen3-32B and EXAONE-3.5-32B 21.2 to 24.7; with a relation replaced by friend, neighbour or coworker, the LLMs answer 3.9 to 74.0 percent. A gate that abstains when the queried people are disconnected nearly matches the derivability check on deletions but catches no replaced relation; the learned derivability check, without rules, is within 0.7 points of the rules. Gating on the derivability check raises the balanced score by 1 to 31 points with a fact deleted (Table~\ref{tab:faith}) and by 2 to 37 with a relation replaced. Gating per condition, the consistency check's reports and the repair of contradicted contexts: the supplementary material.

\tabfaith
\section{The learned consistency check}
\label{app:learned}

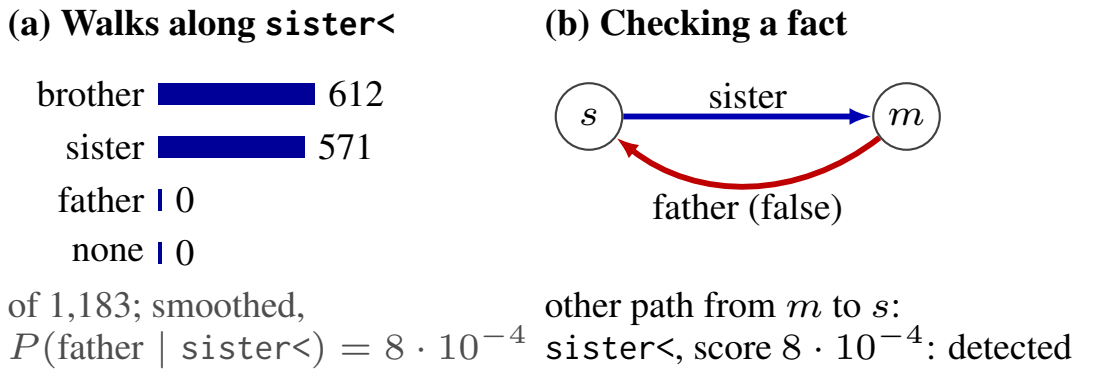
\begin{figure}[t]
\centering
\begin{tikzpicture}[
  font=\scriptsize,
  head/.style={font=\scriptsize\bfseries, anchor=west, inner sep=0pt},
  ent/.style={circle, draw=black!75, fill=white, minimum size=4.4mm, inner sep=0pt},
  sel/.style={-{Latex[length=1.5mm]}, line width=1.1pt, draw=blue!70!black},
  bad/.style={-{Latex[length=1.5mm]}, line width=1.1pt, draw=red!75!black},
  lab/.style={font=\scriptsize, inner sep=1pt},
]
\node[head] at (0,2.55) {(a) Walks along \texttt{sister<}};
\foreach \y/\name/\w/\v in {2.1/brother/0.517/612, 1.75/sister/0.483/571, 1.4/father/0/{0}, 1.05/none/0/{0}} {
  \node[anchor=east, lab] at (0.95,\y) {\name};
  \fill[blue!60!black] (1.0,\y-0.07) rectangle ++({max(0.025,\w*2.0)},0.14);
  \node[anchor=west, lab] at ({1.05+max(0.025,\w*2.0)},\y) {\v};
}
\node[anchor=north west, align=left, inner sep=0pt, text=black!70] at (0,0.8) {of 1,183; smoothed,\\$P(\text{father}\mid\texttt{sister<})=8\cdot10^{-4}$};
\node[head] at (3.55,2.55) {(b) Checking a fact};
\node[ent] (s) at (3.85,1.95) {$s$};
\node[ent] (m) at (5.95,1.95) {$m$};
\draw[sel] (s) -- node[lab, above] {sister} (m);
\draw[bad] (m) to[bend left=38] node[lab, below] {father (false)} (s);
\node[anchor=north west, align=left, inner sep=0pt] at (3.55,0.8) {other path from $m$ to $s$:\\\texttt{sister<}, score $8\cdot10^{-4}$: detected};
\end{tikzpicture}
\caption{The learned consistency check (Appendix~\ref{app:learned}) on Figure~\ref{fig:concept}'s example. (a) Relations the graph
stores between the ends of training walks along one path (\texttt{<}: against a stored fact). (b) The false fact's score from the
other path, below every consistent calibration context's; the GRU addition gives $2\cdot10^{-6}$.}
\label{fig:learned}
\end{figure}

\paragraph{The idea in brief.} \looseness=-1 A graph without rules still shows which relations hold together. In CLUTRR's generated families, for instance, a person's sister's parent is that person's parent, and two people joined by a sister fact are siblings, never parent and child. The learned check collects such regularities by walking the graph at random and counting, for each sequence of relations a walk follows, which relation the graph stores between the walk's two ends. A fact of a context is then suspicious when the other facts of the context join its two entities by a path at whose ends the graph almost never stores the fact's relation: a context in which $m$ is the sister of $s$ and $s$ is the father of $m$ is such a case (Figure~\ref{fig:learned}). The counts come from the graph's own facts and the threshold from contexts known to be consistent, so the check needs no examples of contradictions, which a new graph would not provide. Its limit is that it can judge a fact by path evidence only where such a second path exists; the single-value rule below needs none.

\looseness=-1 Most graphs lack rules, but their facts show how relations constrain each other. The check's three signals mirror three kinds of ontology constraint: relations that cannot hold together, chains of relations that imply another, and relations with a single value; others, such as type mismatches, escape them. \emph{Co-stored relations.} Random walks through the graph, each labelled with the relation the graph stores between its ends or none, give for each relation path $\pi$, its labels read with their directions, counts of the relations stored between its ends (Figure~\ref{fig:learned}a); add-one smoothing turns these into $P(r\mid\pi)$. As with rules, a fact $(s,p,o)$ is compared with what the other facts imply: each path $\pi$ they form between $s$ and $o$, not the fact, scores it by $P(p\mid\pi,\text{stored})=P(p\mid\pi)/(1-P(\text{none}\mid\pi))$, the probability of the claimed relation with the mass of none removed, as graphs often omit relations that hold. In Figure~\ref{fig:learned}b a false fact makes $s$ the father of $m$ while a true fact makes $m$ the sister of $s$; no walk along that path, \texttt{sister<} (the sister relation read against its stored direction), ends at a father, so the false fact scores $8\cdot10^{-4}$. One conflicting path suffices: a fact keeps its lowest score over its paths; paths never walked, or with $P(\text{none}\mid\pi) \geq 0.95$, are skipped. \emph{Composed relations.} Some contradictions join two entities with no stored relation: PathQuestion's knowledge bases store no grandparents, so the counts skip a path that makes $g$ a grandparent of $c$ while a fact makes $c$ a parent of $g$. TransE \citep{bordes2013} gives relations vectors with $h+r \approx t$ for every fact $(h,r,t)$, so a path implies the sum of $+r$ per step along a fact and $-r$ per step against one. A fact conflicts when its relation's vector is far from this sum for another path between its entities: from $g$ to $c$, $-2r_{\text{parent}}$ against the claimed $+r_{\text{parent}}$. \emph{Single values.} A relation with one object for at least 99 percent of its subjects in the graph, such as place of birth, is contradicted by a second object.

\looseness=-1 A context's count and vector scores are those of its most suspicious fact, as one contradicted fact suffices, and each is read as the share of consistent calibration contexts of similar size that score at least as suspiciously. The check detects a contradiction when the smaller share falls below that of 99 percent of these contexts, so that 1 percent of them raise a false alarm, or when a single-valued relation has two objects; the rule was chosen on 1,894 reversed family facts in compact contexts of PathQuestion training questions, and no contradicted example is used to count, to fit the vectors or to set the thresholds, as a new graph would offer none; a path through an unnamed Freebase node that groups the facts of an event, such as a marriage, is read as one relation (below). We walk families from CLUTRR's generator, not test stories, PathQuestion's knowledge bases and 1,200 WebQSP training graphs.

\looseness=-1 \emph{A path classifier as an addition.} Counts know only paths walked in training. A two-layer gated recurrent unit network (GRU; \citealp{cho2014}) trained on the same walks scores any path, as a third share, and serves as CLUTRR's learned derivability check, passing a context when it names a relation for a path between the queried people; it adds detection almost only where a false fact relates CLUTRR people three or more facts apart (far contradictions, and shortcuts when the whole story is checked), so counts are the default, chosen post hoc (below). Table~\ref{tab:learned} gives the check's detection and false-alarm rates, and the rest of this appendix its settings, its limits and the natural errors.

\tablearned

The default consistency check scores facts by smoothed path counts, TransE relation vectors (vectors below) and single values (above); the GRU addition adds the score of a path classifier, a GRU, as a third share, and the GRU check, for comparison, replaces the counts by it.

\paragraph{Training.} \looseness=-1 The counts come from 40,000 random walks of each length (60,000 on WebQSP), of up to 8 facts on CLUTRR and 4 on the knowledge bases (at most 3 to test shorter walks, as many in total), each labelled with a relation stored between its ends, or none, and read with directions (``father>'' along a stored fact, ``father<'' against it). The walks cover 200 families from CLUTRR's generator with the benchmark's settings (3 generations, 4 children per couple, everyone married), which store every related pair's relation in both directions and are never test stories, the three PathQuestion knowledge bases and the first 1,200 WebQSP training graphs. TransE is trained, and single-valued relations counted, on the same graphs; only its relation vectors are kept. The count, vector and GRU scores take 0.20, 0.15 and 0.62 ms per WebQSP compact context (the classifier, the count tables and the timings: supplementary material).

\paragraph{Scoring and calibration.} \looseness=-1 Each fact is scored against up to 64 simple paths that the other facts form between its entities, of up to the training length in contexts of up to 20 facts and of at most 2 facts on CLUTRR and 1 on the knowledge bases in larger ones; its vector score is the largest distance between its relation vector and a path's vector. The shares above are taken within each size bin (up to 20, 21 to 60, more than 60 facts); in the GRU addition the GRU's $P(p\mid\pi,\text{stored})$ gives a third share. One threshold on the smallest share keeps false alarms at 1 percent; 1 percent for each of the default's two scores would allow up to 2. Calibration contexts are held-out pipeline contexts or consistent subgraphs of held-out graphs (per benchmark and bin: supplementary material). On them the default raises 0.0 to 1.2 percent false alarms per bin, below 1 where scores tie at the threshold and above where the single-value rule fires. No contradicted example is used to count, train or calibrate, but smoothed counts became the default after an ablation on the test probes (supplementary material), so that choice is post hoc.

\paragraph{Unnamed records.} \looseness=-1 A path $x \xrightarrow{r_1} v \xrightarrow{r_2} y$ through an unnamed node $v$ (a Freebase record grouping the facts of an event, such as $x \xrightarrow{\text{spouse\_s}} v \xrightarrow{\text{spouse}} y$ through a marriage $v$) is read as one relation $r_1|r_2$ between $x$ and $y$ in training and scoring, and a fact touching an unnamed node serves as evidence and is judged by the single-value rule alone.

\paragraph{Learned derivability on CLUTRR.} \looseness=-1 CLUTRR's learned derivability check passes a context when the GRU names a relation for a path between the queried people, read whole with $P(\text{none})<0.5$ or composed step by step, each step turning the current relation distribution into that of a two-fact path; the learned answer combines both readings. Trained on the default walks, the GRU passes 99.7 percent of clean compact contexts (the rules 99.8) and answers 99.0 to 99.3 percent of chains correctly, with 35 errors. Read the same way, smoothed counts from the default walks pass only 66.0 percent of clean compact contexts, so the derivability check keeps the GRU.

\paragraph{Large contexts.} \looseness=-1 Larger contexts hold more rare but true relation combinations (supplementary material). There the default still detects 96.4 to 99.9 percent of CLUTRR reversed edges but 75.2 to 88.5 of far contradictions, and in SubgraphRAG's 100 WebQSP facts all 33 second values but only 1.2 percent of the 165 contexts with a reversed fact, as many as without it.

\paragraph{Where the GRU adds.} \looseness=-1 On compact contexts the GRU addition raises detection only on CLUTRR far contradictions, from 87.6 to 99.9 percent (the rules 99.8; and on 4 of the 3,675 substitutions entered before selection, supplementary material): it detects 536 that the default misses (12.3 percent of the 4,361), all with the false fact's two people at least 3 facts apart on the other facts. On PathQuestion and WebQSP it only adds false alarms (Table~\ref{tab:learned}): their probes are caught by statistics of single facts, one-fact paths and single values.

\paragraph{Behind the checks.} \looseness=-1 A contradicted compact context stays derivable: with the rules, 4,356 of CLUTRR's 4,361 reversed-edge contexts (99.9 percent) pass the derivability check, so the consistency check is what flags them.

\paragraph{Contamination before selection.} \looseness=-1 The supplementary material tabulates what becomes of a false fact entered into the story, knowledge base or question graph before selection (not selected, insufficient, detected, a wrong or a right answer returned). A shortcut relates the queried people directly by a wrong relation; a substitution replaces the relation (CLUTRR) or object (PathQuestion, WebQSP) of one compact-context fact by a wrong one; on WebQSP a reversed family or containment fact or a second value of a single-valued relation also enters the graph. A substitution yields an undetected wrong answer for 97.9, 41.1 and 43.1 percent of the CLUTRR, PathQuestion and WebQSP questions given one. A reversed WebQSP fact is never selected.

\paragraph{Parts of the check.} \looseness=-1 The ablation (supplementary material) calibrates each variant of the check to 1 percent false alarms. Unsmoothed, a relation that a walked path never ended in scores 0 however few walks the path had; add-one smoothing makes it fall with the number of walks, lifting detection of CLUTRR's far contradictions from 69.1 to 87.6 percent and PathQuestion's reversed family facts from 0.7 to 100.0. On WebQSP all variants detect the same probes.

\paragraph{Natural errors.} \looseness=-1 We also tested WebQSP's check, unchanged on CWQ, on the question graphs' own errors (supplementary material). It has path evidence for only 2.4 and 2.6 percent of named fact mentions in their compact contexts, where an estimated 1.7 percent [0.4, 3.6] of named facts Wikidata covers are wrong if those it supports are right (brackets: 95 percent intervals). Of the 18 errors confirmed there, 2 (one a merge) have a second path, in 3 contexts; the check flags both, in 2 of them, but only with 7 facts judged right at the same score, and no other of its 113 flagged named facts is judged wrong (unflagged ones there: 3 of 78). In denser contexts around 181 rule-contradicted facts it flags 91.7 percent, and 94.4 of supported ones: `inconclusive' by the rule fixed in advance, the sign depending on the labels; its smallest share does not separate the 25 facts adjudicated Freebase wrong from supported facts (AUC 0.52 [0.47, 0.55]), and a threshold re-estimated post hoc flags none. Whether the check detects natural errors thus remains open, since it could judge only 2 of them, and outdated evidence is not covered. The labels are LLM judgements against Wikidata, not the authors': rules that compare a fact with Wikidata's statements for the same property mark it supported or contradicted, and two LLM adjudication passes judge the rest, consulting Wikipedia, with an LLM tie-break (kappa 0.76); the first pass was made by the same LLM that drafted the rules, and two rules were amended, one after labelling began; an independent re-judgment by GPT-5 (below) changes the error count but not the conclusion.

\tabcheckable
\paragraph{Where the check has evidence.}\label{app:checkable} \looseness=-1 The consistency check judges a fact only where the context holds another path between the fact's entities (a count score for an informative path, or a vector score above 0); we call such facts and contexts judgeable by a second path, judgeable for short, and report the false-alarm and detection rates on them, where the rates on all facts would mostly measure how rarely the check can judge anything (the supplementary material tabulates them; brackets: 95 percent Wilson intervals). In the compact WebQSP and CWQ test contexts, 451 of 17,881 named fact mentions are judgeable (2.5 percent; 2.4 and 2.6 per benchmark), in 107 of 2,895 contexts. Compact contexts are sparse by design: some fact has a second path in 2.6 percent of the WebQSP compact contexts (43 of 1,628), 3.0 of CLUTRR's (133 of 4,361) and 26.9 of PathQuestion's (210 of 782), but in 37.7 percent of whole CLUTRR stories, 42.2 of PathQuestion's one-hop neighbourhoods and 92.7 of SubgraphRAG's 100 WebQSP facts (1,509 of 1,628); in these clean CLUTRR and PathQuestion contexts the check flags no compact context and at most 4.2 percent of the others with a second path. There the threshold flags 27.6 percent [23.7, 31.9] of the mentions that are not confirmed errors (31.9 and 25.6), against 27.0 percent [15.4, 43.0] of the 37 judgeable WebQSP calibration contexts (2.7 percent of 1,369), on which the 1 percent overall rate was set: the calibration bound holds overall and is not a bound where the check has evidence. Of the 32 mentions of the confirmed natural errors, 2 are judgeable, and both are flagged (100.0 percent [34.2, 100.0] given evidence, 6.2 [1.7, 20.1] of all error mentions).
\paragraph{Error-centred contexts.} \looseness=-1 In the denser error-centred contexts of the natural-error study (at most 20 facts around the fact under test), 96.0 percent [80.5, 99.3] of the 25 confirmed errors and 94.9 of the 905 supported controls are judgeable, and the check flags 100.0 percent [86.2, 100.0] of the judgeable errors but also 99.4 [98.6, 99.8] of the judgeable controls (98.2 of the 181 rule-contradicted cases): conditioning on evidence raises detection and false alarms alike, which is the AUC of 0.52 [0.47, 0.55] reported above.
\paragraph{Injected probes and contamination, given evidence.} \looseness=-1 A reversed or far contradiction creates the alternative path by construction, so every injected probe is judgeable, and a second value is caught by the single-value rule without any path (83 WebQSP probes, none with path evidence, 80 detected). A false fact entered before selection is judgeable only if it reaches the context: a reversed WebQSP fact never does (0 of 179), a second value always (100.0 percent; 96.4 of these detected), a substitution almost always (98.4 percent) but is detected for 1.3 percent of those.

\paragraph{What the check can judge.} \looseness=-1 Counting the single-value rule, which needs no second path but fires only when the context holds a second value, the check can judge more than the 451 named mentions with a second path (Table~\ref{tab:checkable}): the rule applies to 408 (2.3 percent; 399 without a second path), so 850 of 17,881 named mentions (4.8 percent) are judgeable by either test, in 406 of 2,895 contexts (14.0 percent). The rule fires on no named mention, so all 126 flags come from the path scores: 27.9 percent of the mentions with a second path (27.6 without the 2 confirmed errors) and 14.8 of the judgeable ones; of the 32 mentions of confirmed natural errors, the 2 judgeable ones have a second path, and both are flagged.
\paragraph{The natural-error labels re-judged.} \looseness=-1 The natural-error labels come from two LLM passes and an LLM tie-break against Wikidata (above). To test them, GPT-5 (reasoning effort low), another model family, judged every adjudicated row (408) blind, with the evidence of the second pass and its own knowledge, and 202 facts the rules found supported, of which it calls 198 acceptable (610 calls, USD 3.54 at list price). It agrees with the consensus on 85.3 percent of the rows (kappa 0.74 [0.68, 0.79]; the two original passes 0.76), confirms 71 of the consensus's 74 errors and is stricter, calling 120 rows errors, mostly rows the consensus found acceptable (29: loose relations such as a city's tourist attractions) or could not decide (15). Under its labels the compact contexts hold 41 errors (91 mentions) instead of 18 (32); of its 7 judgeable error mentions 6 are flagged, each with true facts at the same score, and flagged facts are errors no more often than unflagged ones (5.3 against 11.5 percent). On the rows both agree on, 15 errors (28 mentions) remain; under either labelling the check separates no errors from true facts in the denser contexts (AUC 0.53 and 0.52), and both errors it detected under the consensus labels are disputed (in full: the supplementary material). The conclusion that the check does not separate natural errors from true facts holds under all three labellings; the number of errors and the prevalence estimate (1.7 percent; 4.2 under GPT-5's labels) depend on them, so the labels are judgements, not verified facts.

\section{What a passing check guarantees}
\label{app:layers}

\paragraph{Four layers on WebQSP.} \looseness=-1 A passing check on WebQSP can be read as four claims, which we call layers and test on each test question: (1)~interpretation: the predicted chain and constraints are exactly those of an annotated parse of the question; (2)~execution: on the context's facts alone they leave some answer, every constraint testable; (3)~completeness: that answer set equals the one on the question's whole graph (the question graph); (4)~consistency: the default consistency check finds no conflict. The derivability check passes 743 of the 1,628 questions (45.6 percent). All satisfy layer~2, by construction, but only 589 (79.3 percent) layer~1. Layer~3 fails for 14 of the 34 passing contexts cut at 100 facts; the returned answer, executed on the whole graph, is unaffected. The consistency check flags 4 passing contexts (11 with the GRU check); the derived answer is exact for 3 of the 4 and 10 of the 11, and for 585 of all passing questions (78.7 percent). A pass means that the answer is derived under the predicted query, not that it is verified.

\paragraph{The first layer that fails.} \looseness=-1 The supplementary material assigns each question to its first failing layer; the counts are chain not annotated 419, constraint missed 144, other constraints 82, no execution 65, incomplete 8, conflict found 8, 1 to 4 hold, wrong 63 and all five hold 839, where the group `1 to 4 hold, wrong' passes all four layers with a wrong answer and the last group with a right one.

\paragraph{Wrong answers that pass.} \looseness=-1 Of the 158 wrong derived answers among passing questions (one withheld by the consistency check), 55 lie on an unannotated chain and 103 on an annotated one: 69 supersets of the gold set, 63 of them with a constraint of the annotated parse that the neighbours' vote did not apply, 38 of these a date, `now' or ordering constraint the graph cannot test; 21 subsets; 7 overlapping and 6 disjoint.

\paragraph{Checking the context against the graph.} \looseness=-1 A graph-aware check would also run the predicted query on the whole graph and pass only where the context yields the same non-empty answer set. On depleted contexts it passes 0, 56 and 0 of the 777, 801 and 1,438 PathQuestion, MetaQA and WebQSP contexts where the derivability check passes 80, 404 and 419, and gating on it raises every LLM's balanced score further (the supplementary material); on the selector's own contexts it costs nothing.

\paragraph{Right answers where no answer follows.} \looseness=-1 On the 636 depleted WebQSP contexts from which no answer follows, the seven open LLMs under the abstention prompt abstain on 56.8 to 74.5 percent, answer wrongly on 11.6 to 34.0, rightly with every gold answer named in the context (so readable from facts off the predicted chain) on 6.0 to 15.3, and rightly from own knowledge (a gold answer not named) on 0.0 to 4.7; 21.5 to 57.5 percent of their answers are right.

\paragraph{Where the derived answer is executed.}\label{app:executed} \looseness=-1 On CLUTRR, PathQuestion and MetaQA the check and the derived answer are executed on the final context's facts only; the selector built that context from the graph, so it holds every path of the predicted sequence, and completeness rests on the selector (Section~\ref{sec:method}). On WebQSP and CWQ the selector enumerates the predicted chain's paths in the whole question graph, tests the constraints there and returns the path ends, while the context it writes is cut at 100 facts: the derived answer of Tables~\ref{tab:answer} and~\ref{tab:answerfull}, of the layer tables (supplementary material) and of Appendix~\ref{app:cwq} is the graph execution. The sentence of Section~\ref{sec:method} that the check reads only the context describes the check applied to a given context. Executed on the context only, the derived answer where the check passes falls from 78.7 to 78.1 percent on WebQSP and from 52.0 to 51.8 on CWQ; routing moves by $-$0.3 points for every open LLM and for GPT-5 on WebQSP and not at all on CWQ; the pass set is unchanged on WebQSP and would lose 2 CWQ questions. The graph execution is kept because the selector has already run the chain on the graph (no further cost) and because the context cut is a presentation limit for the LLM, not a limit of the evidence the selector saw.

\paragraph{Literal constraints and unreachable answers.} \looseness=-1 RoG's question graphs keep no literals, so a date, `now', numeric, ordering or literal-match constraint of the gold query cannot be tested (classified by the gold SPARQL; on WebQSP any of a question's parses). Such questions are 10.3 percent of WebQSP and 28.6 of CWQ; they pass the check 35.1 and 6.4 percent of the time, with 22.0 and 3.1 percent of the derived answers exact, and hold 46 of the 158 wrong passing answers on WebQSP (among them the 38 untestable constraints counted above) and 63 of 200 on CWQ, where no comparative (7 passing) or superlative (6) question passes with a right answer. The graph reaches no gold answer for 71 WebQSP and 683 CWQ questions (4.4 and 19.3 percent). On questions it can test, reachable and free of literal constraints (86.1 and 55.7 percent), the check passes 48.5 and 17.9 percent, with 84.1 [81.2, 86.7] and 61.1 [55.9, 66.0] percent exact, against 78.7 and 52.0 on all: literals and unreachable answers explain a minority of the wrong passing answers (50 of 158 and 63 of 200); the rest (108 and 137) is mostly the method's (chain, constraint, answer set), with other graph gaps such as missing or renamed answers.
\paragraph{WebQSP's routing gate.} \looseness=-1 Executed on the final context only, the derived answer where the routing gate passes falls from 65.1 to 64.8 percent (where it fails from 6.2 to 5.8; on all questions from 56.3 to 56.0), the answer sets differ for 16 of the 1,385 passing questions, and routing moves by $-$0.3 points for every open LLM.

\section{Who answers: derived answer or LLM}
\label{app:routing}

\looseness=-1 Table~\ref{tab:gates} collects the coverage and precision of the check and of both gates on test and held-out questions. Table~\ref{tab:answerfull} extends Table~\ref{tab:answer} with the accuracy where the routing gate fails, the gain of routing over the LLM alone and the number of LLMs whose routing beats the derived answer. On WebQSP, behind its routing gate, routing beats the four smallest LLMs alone by 6.7 to 12.5 points (Holm-adjusted exact McNemar $p \leq 5.8\times10^{-12}$), changes the three largest by $-$0.7 to +0.6 ($p = 1$) and costs GPT-5 2.7 and 4.0 points without and with reasoning ($p = 6.3\times10^{-4}$, $1.9\times10^{-8}$).

\tabgates
\tabanswerfull

\paragraph{What the WebQSP check adds.} \looseness=-1 With the chain predictor, contexts and derived answers fixed, the supplementary material adds the check's parts one at a time, on clean test questions and contradicted probes. A vote threshold alone, set to return as many derived answers, leaves 23.8 rather than 21.3 percent of them wrong (95 percent bootstrap interval of the difference 1.0 to 4.1 points), and the consistency check cuts wrong derived answers on contradicted probes from 10.1 and 7.2 to 1.7 and 0 percent (reversed fact, second value). GPT-5 reads the 1,385 questions passing WebQSP's routing gate at 68.3 and 69.8 percent without and with reasoning, against the derived answer's 65.1, so routing scores 60.0 and 60.4 percent of all questions, against 62.7 and 64.4 for GPT-5 alone (exact McNemar $p = 6.3\times10^{-4}$ and $1.9\times10^{-8}$) and 56.3 for the derived answer alone.

\paragraph{Who answers, chosen on validation questions.} \looseness=-1 WebQSP's 246 validation questions get contexts, gates and derived answers built as the test ones; the routing gate passes for 199, where the derived answer is right for 68.3 percent. Per LLM, validation then chooses among four fixed policies (the LLM alone; routing; the derived answer or an abstention; the LLM where the routing gate passes, else an abstention) by accuracy and by a score of $+1$ per right answer, $-1$ per wrong one and 0 per abstention, every reply counting as an answer (per LLM: the supplementary material).

\paragraph{Confidence estimators for the derived answer.}\label{app:riskcov} \looseness=-1 The derivability check is one confidence estimator for the derived answer: the predicted chain's share of the neighbours' votes, with questions whose applied constraints cannot all be tested never returned, at a threshold fixed on validation questions. We compare it with seven others on the same questions by risk-coverage curves (coverage: the share of questions whose derived answer is returned; risk: the share of returned answers that are not exact): the vote share alone; the margin to the second realisable chain; the nearest-neighbour and the mean voter similarity; fewer distinct voter chains; a smaller derived answer set; and SubgraphRAG's top-1 triple score. Each is read at the smallest coverage at or above the check's (45.6 and 11.8 percent) and by the area under its curve (AURC; all eight: the supplementary material; brackets: percentile bootstrap over questions, 2,000 resamples). On WebQSP the check's risk at its coverage, 21.3 percent [18.0, 24.0], is below every other estimator's (23.8 for the vote share alone, 23.8 to 30.3 for the six others; paired differences of +2.6 to +9.0 points, all intervals above 0), and its AURC over its own coverage range, 21.4, is the lowest (over a common range: below). On CWQ its risk is 48.0 percent [42.8, 53.0]; the vote share alone (70.8 percent), the margin and the number of voter chains are worse by +19.3 to +24.5 points, the two similarities are indistinguishable from it, and two estimators are better: SubgraphRAG's top-1 score (40.8 percent at the same coverage, $-$7.2 [$-$13.4, $-$1.1]) and the size of the derived answer set, whose risk is 31.4 percent at 24.5 percent coverage, the smallest its tie groups allow, as 24.5 percent of CWQ questions have a one-element derived set. Ranked by set size with ties broken by the vote share and cut at the check's coverage, the set size has a risk of 14.1 percent [10.8, 17.6], the figure of Section~\ref{sec:checks}, and an AURC of 47.1 against the check's 49.6. On CWQ, where the chain vote is rarely confident and often wrong, the constraint test and the vote are weaker signals than the answer set's size or the retriever's own score; on WebQSP no estimator beats the check, and SubgraphRAG's top-1 score is indistinguishable from it over the common range (below).
\paragraph{Routing by another estimator.} \looseness=-1 Routing (the derived answer where returned, else the LLM on our context) is insensitive to the estimator on WebQSP: at the check's coverage, the mean exact match of the seven open LLMs under every other estimator is within 1.0 point of routing by the check. On CWQ, routing by the size of the derived answer set, at its 24.5 percent coverage, is +2.9 points [+2.4, +3.5] above routing by the check (mean 31.0 against 28.1); chosen on validation, it is CWQ's routing gate (below). The threshold of the check's own estimator that maximises routed exact match on the test questions (optimistic) would add at most 5.0 points on WebQSP and 6.2 on CWQ over the validation-chosen one (per LLM: supplementary material).
\paragraph{Gating the LLM's own answer.} \looseness=-1 Used instead as a gate on the LLM's own answer (the answer is returned where the estimator is at or above a threshold, the derived answer never used), the check lowers the share of wrong returned answers at its coverage from between 35.6 and 54.7 percent without a gate to between 17.8 and 38.2 on WebQSP (nine LLMs), and from between 61.8 and 84.6 percent to between 39.6 and 75.1 on CWQ (seven); it has the lowest AURC of the eight estimators for 7 of the nine WebQSP LLMs and for 0 of the seven on CWQ.
\paragraph{Choosing the routing gate on validation.} \looseness=-1 The routing gates of WebQSP and CWQ are chosen on validation questions by one procedure, fixed with the gates in a pre-specified plan before the held-out confirmation of Appendix~\ref{app:prereg}: among 43 rules on the check and the size of the derived answer set (the check; set size at most $k$; both; either), the one with the highest mean routed exact match of the seven open LLMs on validation. The candidate family was defined after comparing gates on test questions; the rule then chose the routing gates on validation, and they and the precision gate were confirmed on held-out questions. On CWQ's 3,519 validation questions it picks `set size at most 1' (routed mean 33.8 against 30.5 for the check; test 31.0 against 28.1, +2.9 [+2.4, +3.5]), the one-answer gate of Appendix~\ref{app:cwq}. On WebQSP's 246 it picks `the check, or set size at most 20' (a near tie with `the check, or set size at most 8 to 10'), which passes 80.9 percent of the validation and 85.1 of the test questions with 68.3 and 65.1 percent of the derived answers exact (the check: 45.6 and 78.7 on test). On PathQuestion and MetaQA no validation reads exist, so their routing gate is the check; post hoc, the check is the best rule of the family on their test questions.

\paragraph{A common coverage range.} \looseness=-1 The AURC above averages each estimator's risk over coverage from 0 to its own maximal coverage. The check never returns a question whose applied constraints cannot be tested, so its range ends at 84.4 percent on WebQSP and 42.6 on CWQ, the others' at 88.5 and 67.8 (an empty derived answer is never returned). Over the common range, from 0 to the check's maximal coverage and divided by its length, no estimator has a lower AURC than the check on WebQSP (21.4); SubgraphRAG's top-1 triple score is indistinguishable (21.2, $-$0.3 [$-$2.2, +1.9]). On CWQ the smaller derived answer set is lower (39.2 against 49.6, $-$10.4 [$-$13.7, $-$6.9]), and over the check's operating range, up to its coverage of 11.8 percent, so are three estimators (mean voter similarity, smaller derived answer set, SubgraphRAG's top-1 score). Which estimator gates CWQ is decided on validation questions (Appendix~\ref{app:cwq}).
\paragraph{Six candidate policies on WebQSP.} \looseness=-1 With validation reads under the abstention prompt, two candidates join the four fixed policies (the LLM alone; routing; the derived answer where the routing gate passes, else an abstention; the LLM where the gate passes, else an abstention): the LLM under the abstention prompt alone, and behind the derived answer where the routing gate passes (ties go to fewer LLM calls). By accuracy validation picks routing for six open LLMs and routing to the abstention-prompt LLM for Gemma-3-27B, by the score the derived answer or an abstention for six and the LLM's own abstention for Qwen3-32B; no policy chosen for an open LLM is significantly worse than the LLM alone on its own criterion (held-out: Appendix~\ref{app:prereg}; test: the supplementary material). Validation overstates the derived answer's precision where the routing gate passes: 68.3 percent on validation against 65.1 on test, and on CWQ 69.2 against 68.6 (per LLM, with the regret against the best policy in hindsight: the supplementary material).
\paragraph{The precision gate.} \looseness=-1 The precision gate, the check and one derived answer, was named in the pre-specified plan after its CWQ test precision had been seen, so its held-out precision (Appendix~\ref{app:prereg}) is the confirmatory number. On WebQSP's test questions it returns 309 of 1,628 (19.0 percent) with 91.3 percent [87.6, 93.9] exact, above five open LLMs reading the same questions (Holm-adjusted $p<0.05$), level with Qwen3-32B and EXAONE-3.5-32B (90.3 and 90.3) and with GPT-5 (89.6 and 90.6 without and with reasoning, $p \geq 0.27$); on CWQ's, 194 of 3,531 (5.5 percent) with 96.4 percent exact, above all seven open LLMs and Qwen3-32B with reasoning, and level with or above GPT-5 (Appendix~\ref{app:cwq}). On WebQSP GPT-5's policy is chosen on its own validation reads: with reasoning validation picks GPT-5 alone by accuracy; without it, routing, on a near tie where the routing gate passes (135 against 136 of 199 right), which costs 2.7 points on test (exact McNemar $p = 6.3\times10^{-4}$).

\section{Reasoning and proprietary LLMs}
\label{app:reason}

\looseness=-1 The subset contains 12 questions from each CLUTRR combination of story type and chain length (264), 50 from each PathQuestion subset (150) and 50 from each MetaQA hop subset (150). On the CLUTRR subset, GPT-5 and GPT-5-mini without reasoning, and Qwen3-32B, DeepSeek-R1-Distill-Llama-70B and GPT-5 with it, read all facts of the story about as accurately as the compact context: all facts are read better by 0.8 and 3.8 points by GPT-5 and GPT-5-mini, the compact context by 1.1 to 2.3 by the three with reasoning, with $p \geq 0.05$, and $p \geq 0.14$ without GPT-5-mini. Per model, with tokens, the reasoning traces and an LLM exploring the graph like Think-on-Graph: the supplementary material.

\paragraph{All PathQuestion and MetaQA questions.} \looseness=-1 Table~\ref{app:reason:full} repeats Tables~\ref{tab:inputs} and~\ref{tab:answer} on all 782 PathQuestion and 900 MetaQA questions for GPT-5 (effort low) and Qwen3-32B, both with reasoning. GPT-5 with reasoning reads our compact context more accurately in exact match than SubgraphRAG's facts (91.2 against 86.2 percent on PathQuestion, 75.6 against 67.0 on MetaQA) and than the one-hop neighbourhood (77.0 and 57.6; all four Holm-adjusted $p \leq 7.5\times10^{-6}$). Where the derivability check passes, the derived answer is right more often than GPT-5 with reasoning for all three kinds of relation knowledge (96.2 against 91.9 percent with PathQuestion's lexicon, 99.9 against 97.0 with MetaQA's classifier, 99.6 against 98.0 with MetaQA's lexicon), and routing raises the exact match of GPT-5 with reasoning alone by 3.3, 2.9 and 1.2 points in the same order (all six Holm-adjusted $p \leq 9.8\times10^{-4}$). For Qwen3-32B with reasoning, our compact context scores 82.2 and 73.4 percent in exact match on PathQuestion and MetaQA, against 75.3 and 60.6 for SubgraphRAG's facts and 68.7 and 53.3 for the one-hop neighbourhood (Holm-adjusted $p \leq 1.8\times10^{-6}$), and where the check passes the derived answer scores 96.2, 99.9 and 99.6 against its 83.1, 95.0 and 95.2 (Holm-adjusted $p \leq 3.8\times10^{-9}$).

\begin{table*}[!t]
\centering
\scriptsize
\setlength{\tabcolsep}{4pt}
\resizebox{\ifdim\width>\textwidth \textwidth\else \width\fi}{!}{%
\begin{tabular}{l rrr rrr rrr rrr rrr rrr}
\toprule
\textbf{(a)} contexts of Table~\ref{tab:inputs} & \multicolumn{9}{c}{\textbf{PathQuestion, 782 questions}} & \multicolumn{9}{c}{\textbf{MetaQA, 900 questions}} \\
\cmidrule(lr){2-10}\cmidrule(lr){11-19}
 & \multicolumn{3}{c}{GPT-5} & \multicolumn{3}{c}{Qwen3-32B} & \multicolumn{3}{c}{open LLMs} & \multicolumn{3}{c}{GPT-5} & \multicolumn{3}{c}{Qwen3-32B} & \multicolumn{3}{c}{open LLMs} \\
\cmidrule(lr){2-4}\cmidrule(lr){5-7}\cmidrule(lr){8-10}\cmidrule(lr){11-13}\cmidrule(lr){14-16}\cmidrule(lr){17-19}
\textbf{context} & exact & Hit & F1 & exact & Hit & F1 & exact & Hit & F1 & exact & Hit & F1 & exact & Hit & F1 & exact & Hit & F1 \\
\midrule
one-hop neighbourhood & 77.0 & 79.9 & 78.8 & 68.7 & 72.6 & 71.1 & 50.2 & 70.2 & 62.2 & 57.6 & 78.2 & 64.6 & 53.3 & 73.7 & 60.9 & 47.3 & 75.2 & 59.3 \\
similarity top-$k$, as many facts as ours & 38.6 & 49.1 & 46.9 & 35.2 & 45.7 & 42.6 & 33.0 & 51.9 & 44.8 & 24.9 & 50.4 & 35.9 & 24.6 & 47.3 & 35.0 & 26.0 & 56.5 & 39.6 \\
KAPING & 19.3 & 27.4 & 21.6 & 20.5 & 29.3 & 23.1 & 19.0 & 33.4 & 26.2 & 32.6 & 45.4 & 36.6 & 31.1 & 41.4 & 34.8 & 31.3 & 50.9 & 38.7 \\
SubgraphRAG & 86.2 & 90.0 & 88.6 & 75.3 & 80.3 & 78.4 & 52.4 & 70.7 & 63.5 & 67.0 & 91.3 & 77.5 & 60.6 & 87.7 & 73.9 & 49.7 & 86.0 & 66.9 \\
Think-on-Graph$^{a}$ & 55.8 & 69.8 & 61.5 & 50.3 & 65.6 & 56.6 & 44.5 & 65.8 & 55.8 & 36.0 & 57.4 & 44.9 & 35.4 & 54.7 & 43.7 & 33.5 & 59.8 & 44.4 \\
selected compact context (ours)$^{b}$ & 91.2 & 93.4 & 92.5 & 82.2 & 84.1 & 83.5 & 72.4 & 87.7 & 82.2 & 75.6 & 78.3 & 76.6 & 73.4 & 77.9 & 76.4 & 59.5 & 79.8 & 71.1 \\
\midrule
annotated path (reference) & 72.8 & 95.3 & 82.6 & 65.1 & 86.6 & 74.3 & 59.3 & 90.4 & 74.5 & 97.0 & 99.0 & 98.8 & 95.0 & 99.1 & 98.8 & 72.4 & 96.6 & 87.9 \\
compact context, classifier$^{c}$ & -- & -- & -- & -- & -- & -- & -- & -- & -- & 97.0 & 99.0 & 98.8 & 95.0 & 99.1 & 98.8 & 72.4 & 96.5 & 87.9 \\
\bottomrule
\end{tabular}
}%
\par\medskip
\resizebox{\ifdim\width>\textwidth \textwidth\else \width\fi}{!}{%
\begin{tabular}{l l r r r r r r r r r}
\toprule
\multicolumn{3}{l}{\textbf{(b)} who answers (Table~\ref{tab:answer}), exact match} & \multicolumn{3}{c}{\textbf{check passes}} & \textbf{check fails} & \multicolumn{3}{c}{\textbf{all questions}} & \textbf{check passes} \\
\cmidrule(lr){4-6}\cmidrule(lr){7-7}\cmidrule(lr){8-10}\cmidrule(lr){11-11}
\textbf{relation knowledge} & \textbf{LLM} & \textbf{passes} & \textbf{derived} & \textbf{LLM} & \textbf{derived $-$ LLM ($p$)} & \textbf{LLM} & \textbf{LLM alone} & \textbf{routed} & \textbf{routed $-$ LLM ($p$)} & \textbf{best open LLM} \\
\midrule
PathQuestion, lexicon & GPT-5 & 602 of 782 & 96.2 & 91.9 & +4.3 (1.1$\mathrm{e}{-04}$) & 88.9 & 91.2 & 94.5 & +3.3 (1.1$\mathrm{e}{-04}$) & 86.2 \\
PathQuestion, lexicon & Qwen3-32B & 602 of 782 & 96.2 & 83.1 & +13.1 (1.3$\mathrm{e}{-17}$) & 79.4 & 82.2 & 92.3 & +10.1 (1.3$\mathrm{e}{-17}$) & 86.2 \\
MetaQA, classifier & GPT-5 & 900 of 900 & 99.9 & 97.0 & +2.9 (3.0$\mathrm{e}{-08}$) & -- & 97.0 & 99.9 & +2.9 (3.0$\mathrm{e}{-08}$) & 89.0 \\
MetaQA, classifier & Qwen3-32B & 900 of 900 & 99.9 & 95.0 & +4.9 (2.3$\mathrm{e}{-13}$) & -- & 95.0 & 99.9 & +4.9 (2.3$\mathrm{e}{-13}$) & 89.0 \\
MetaQA, lexicon & GPT-5 & 694 of 900 & 99.6 & 98.0 & +1.6 (9.8$\mathrm{e}{-04}$) & 0.0 & 75.6 & 76.8 & +1.2 (9.8$\mathrm{e}{-04}$) & 90.6 \\
MetaQA, lexicon & Qwen3-32B & 694 of 900 & 99.6 & 95.2 & +4.3 (3.7$\mathrm{e}{-09}$) & 0.0 & 73.4 & 76.8 & +3.3 (3.7$\mathrm{e}{-09}$) & 90.6 \\
\bottomrule
\end{tabular}
}%
\caption{All 782 PathQuestion and 900 MetaQA questions, forced-choice prompt (\%). GPT-5 (reasoning effort low) and Qwen3-32B read with reasoning; open LLMs: mean of the seven answering directly. (a) The contexts of Table~\ref{tab:inputs}, for the published methods the facts each selects, not its full pipeline; exact: whole gold set; Hit: some gold answer. $^{a}$Facts Think-on-Graph collected with Qwen3-32B without reasoning; open LLMs: own exploration. $^{b}$The lexicon's context. $^{c}$Same prompts as the annotated path; open LLMs read them in separate runs (Appendix~\ref{app:setup}: up to 1.0 point apart). --: no PathQuestion classifier. (b) Split by the derivability check, the routing gate on these benchmarks; passes: questions passing it; derived: the derived answer; LLM: the LLM on the compact context; routed: the derived answer if the check passes, else the LLM; best open LLM: of the seven where the check passes; --: no question fails. Differences in points, computed before rounding; $p$: exact McNemar, Holm-adjusted over the two reasoning LLMs.}
\label{app:reason:full}
\end{table*}

\paragraph{Answers at the output limit.}\label{app:caps} \looseness=-1 The supplementary material gives, per condition, the answers of Table~\ref{app:passfail:split} and of Table~\ref{app:reason:full} that reached the output limit. In Table~\ref{app:reason:full}, 36 of GPT-5's 12,674 answers and 271 of Qwen3-32B's are capped; the largest share in a column is 5.4 percent (PathQuestion, Qwen3-32B, reasoning, similarity top-$k$), GPT-5's largest 0.9 (PathQuestion, one-hop neighbourhood).
\paragraph{What a capped answer could change.} \looseness=-1 Two bounds cover every outcome of a capped answer, the worst case against the observed sign and the paired drop of questions with a capped answer; under them 13 and 14 of Table~\ref{app:reason:full}'s 14 significant LLM contrasts and 10 and 12 of its 12 derived-answer and routing cells keep $p<0.05$, 2 of the 34 significant cells of Table~\ref{app:passfail:split}'s settings with reasoning LLMs or GPT-5, passing and failing sides, lose significance under the worst case, and no sign reverses (in full: the supplementary material).
\paragraph{Re-asked with a larger limit.} \looseness=-1 Every capped GPT-5 answer was re-asked once with a limit of 8,192 tokens, same prompt, message format and reasoning effort (a new sample, not a continuation): all 193 finished under the new limit, 74 of them now exact, for USD 2.50 at list price. With the rerun answers substituted, the largest movement of a difference in Table~\ref{app:reason:full} is 0.8 points and in Table~\ref{app:passfail:split} 1.3; no GPT-5 cell changes sign or significance. Qwen3-32B's 271 capped answers in Table~\ref{app:reason:full} were re-asked the same way with a limit of 8,192 tokens in a window of 16,384: 197 finished, 62 of them now exact. With these answers substituted, its largest movement of a difference is 0.8 points in Table~\ref{app:reason:full} and 0.6 in Table~\ref{app:passfail:split}; no Qwen3-32B cell gains or loses significance and no significant one changes sign, also when the 74 still capped count at the worst case.

\paragraph{GPT-5 on all WebQSP questions.} \looseness=-1 Without reasoning GPT-5's exact match is higher on our context than on SubgraphRAG's 100 facts (62.7 and 57.9 percent, $p = 1.4\times10^{-4}$), with reasoning similar (64.4 and 65.0, $p = 0.63$; Hit 77.9 and 85.4, F1 72.4 and 76.8), at 380 and 2,019 prompt tokens.

\paragraph{Reporting a contradiction.} \looseness=-1 Under the status prompt, which also offers the reply that the facts conflict, the seven open LLMs report at most 18 percent of contradicted contexts on all CLUTRR questions and 50 percent on PathQuestion, whose contradiction is a reversed family fact, and call up to 22 and 30 percent of sufficient, consistent ones contradictory (on CLUTRR 7 percent without distractors). Qwen3-32B, DeepSeek-R1-Distill-Llama-70B and GPT-5 with reasoning report 47 to 60 percent on PathQuestion with at most 2.0 percent false alarms, and on the CLUTRR subset 97.0, 89.4 and 89.4 percent of reversed-edge and 87.5, 86.4 and 57.2 percent of far contradictions, with at most 6.4 percent false alarms (DeepSeek-R1-Distill-Llama-70B with eight consistent distractors), at 703, 722 and 542 output tokens per reply, against 4 and 13 for Qwen3-32B answering directly and GPT-5 without reasoning (on PathQuestion 474, 355 and 253 against 10 and 19); the supplementary material gives these runs per LLM, GPT-5 on all WebQSP questions and the reasoning traces that mention a conflict.

\paragraph{Depleted contexts on knowledge bases.} \looseness=-1 Deleting one fact of a PathQuestion or MetaQA context leaves no path to an answer in 41 and 44 percent of cases; there the seven open LLMs still answer 17.4 to 50.6 and 2.0 to 25.9 percent of the contexts (in full: the supplementary material).

\paragraph{Abstaining on sufficient contexts.} \looseness=-1 GPT-5 with reasoning abstains on 74.0 percent of the sufficient CLUTRR subset contexts real kinship leaves ambiguous and on 32.3 percent of the unambiguous ones (without reasoning 38.9 and 17.3; Qwen3-32B and DeepSeek-R1-Distill-Llama-70B with reasoning at most 10.7). It applies real kinship, which often leaves CLUTRR's label open (Appendix~\ref{app:depth}).

\paragraph{The rules given to the LLM.} \looseness=-1 To separate this reading from a failure to compose, GPT-5 with reasoning rereads the CLUTRR subset under the status prompt with CLUTRR's conventions and its 16 composition rules before the reply options (the supplementary material). It then calls 9.8 rather than 52.7 percent of the consistent contexts insufficient (1.5 rather than 36.1 percent of those without a step or in-law composition), answers 89.0 rather than 44.7 percent correctly and reports 78.4 rather than 57.2 percent of far contradictions, still calling 98.5 percent of depleted contexts insufficient. The reasoning LLMs' remaining gap lies on questions real kinship leaves open: on the 133 subset questions free of step and in-law compositions, Qwen3-32B, DeepSeek-R1-Distill-Llama-70B and GPT-5 with reasoning (forced-choice prompt) are right for 100.0, 99.2 and 99.2 percent, as is the derived answer (100.0), and on the other 131 for 85.5, 84.0 and 84.0 against 98.5 ($p \leq 1.5\times10^{-5}$).

\paragraph{WebQSP under the abstention prompt.} \looseness=-1 Deleting one fact on the predicted chain's paths (1,531 of 1,628 contexts) leaves no answer for 41.5 percent, some for 52.4 and all for 6.1; the check fails every context left with none and passes the others exactly when it passed the full one. Offering abstention costs 2.5 to 5.7 points on compact contexts, where the LLMs abstain on 14 to 23 percent; where no answer follows they answer 25 to 43 percent, and where some or all does, 80 to 93. Behind the check the balanced score rises from 0.48 to 0.66 to 0.63 to 0.67 (per LLM and a corrected score: the supplementary material; split: Appendix~\ref{app:layers}).

\section{Held-out confirmation under a pre-specified plan}
\label{app:prereg}

\paragraph{Design.} \looseness=-1 The gate rule of Appendix~\ref{app:riskcov}, the resulting routing gates and the precision gate, four primary hypotheses with their thresholds and tests, and the secondary analyses were fixed in a pre-specified plan, a timestamped commit to the code repository on 2026-10-07 (hash and plan file released with the code), before any held-out question was selected; the analysis script was frozen, its hash recorded, before any held-out read. The held-out questions come from the training splits and were never used for evaluation: 500 WebQSP questions drawn at random and 1,501 CWQ questions of 168 whole seed clusters (the WebQSP question a CWQ question was composed from), as the CWQ test set is built; they and their near-duplicates were removed from the chain-voting pool, and the seven open LLMs read our context for each under the forced-choice and the abstention prompt. WebQSP questions are the units of every test; on CWQ seeds are (seed bootstrap, seed sign-flip test).
\paragraph{Primary hypotheses.} \looseness=-1 H1, the chosen routing gate's precision passes the pre-set tolerance: WebQSP 66.0 percent (282 of 427) [61.4, 70.4] against 68.3 on validation, CWQ 69.0 (234 of 339) [59.5, 76.6] against 69.2; both at or above the validation Wilson lower end, the tolerance (61.6, 66.0), and above 50 percent (one-sided $p = 1.6\times10^{-11}$; seed-bootstrap 5th percentile 60.9): confirmed. H2, routing beats each LLM alone (Holm over seven): confirmed for the four smaller LLMs on both benchmarks (+6.6 to +8.6 and +4.7 to +9.3 points), not for the three largest ($-$0.6 to +1.4 and $-$0.8 to +0.1); the mean over the seven gains +4.7 [+2.1, +7.4] and +3.4 [+2.0, +4.8]. H3, the chosen routing gate against the check in mean routed exact match: WebQSP +0.3 [$-$1.5, +2.1], not distinguished (per LLM $-$2.6 to +2.4, none significant), CWQ +2.6 [+1.4, +3.9], better; the gates are not changed whatever H3 shows. H4, the precision gate's precision (the check and one derived answer) is above the pre-set 80 percent: WebQSP 89.8 (79 of 88; one-sided lower bound 82.8), CWQ 95.9 (70 of 73; seed-bootstrap 5th percentile 89.5): confirmed; 89.8 and 95.9 are point estimates on 88 and 73 returned answers.
\paragraph{Secondary analyses.} \looseness=-1 With the answerer chosen per open LLM on validation (Appendix~\ref{app:riskcov}): chosen by accuracy, accuracy against the LLM alone changes by +6.6 to +8.6 points for the four smaller LLMs on WebQSP and +4.7 to +9.3 on CWQ and by $-$0.6 to +0.8 and +0.0 to +0.1 for the three larger; chosen by the equal-cost score, the score changes by +4.6 to +26.6 and +31.2 to +74.8 (significantly for 6 and 7 of the seven); no change on a policy's own criterion is significantly negative. The other secondary analyses are in the supplementary material.
\paragraph{Deviations.} \looseness=-1 The supplementary material lists 13 departures from the pre-specified plan, none changing a gate, threshold, LLM or decoding setting.

\end{document}